\documentclass{article}

\usepackage{arxiv}

\usepackage[utf8]{inputenc} 
\usepackage[T1]{fontenc}    
\usepackage{hyperref}       
\usepackage{url}            
\usepackage{booktabs}       
\usepackage{amsfonts}       
\usepackage{nicefrac}       
\usepackage{microtype}      
\usepackage{lipsum}		
\usepackage{graphicx}
\usepackage{doi}
\usepackage[
  separate-uncertainty = true,
  multi-part-units = repeat
]{siunitx}

\usepackage{caption}
\usepackage{subcaption}
\usepackage{booktabs}    
\usepackage{pifont}
\newcommand{\xmark}{\ding{55}}%

\usepackage{adjustbox}
\usepackage{multirow}
\usepackage{tabularx,ragged2e,booktabs}

\title{Exploring Dimensionality Reduction Techniques in Multilingual Transformers }

\date{} 					

\author{ \href{https://orcid.org/
0000-0003-2165-0144}{\includegraphics[scale=0.06]{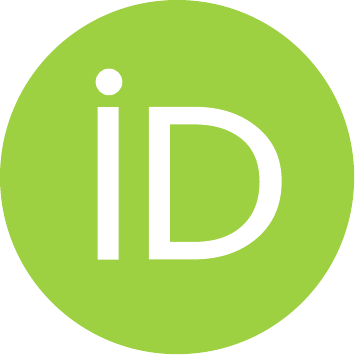}\hspace{1mm}{\'A}lvaro Huertas-Garc{\'i}a}\thanks{Use footnote for providing further
		information about author (webpage, alternative
		address)---\emph{not} for acknowledging funding agencies.} \\
	Computer Systems Department\\
	Universidad Polit{\'e}cnica de Madrid\\
	Madrid, Spain \\
	\texttt{alvaro.huertas.garcia@upm.es} \\
	\And
	\href{https://orcid.org/
0000-0002-0800-7632}{\includegraphics[scale=0.06]{orcid.pdf}\hspace{1mm}Alejandro Mart{\'i}n} \\
	Computer Systems Department\\
	Universidad Polit{\'e}cnica de Madrid\\
	Madrid, Spain \\
	\texttt{alejandro.martin@upm.es} \\
	\And
	\href{https://orcid.org/
0000-0003-4127-5505}{\includegraphics[scale=0.06]{orcid.pdf}\hspace{1mm}Javier Huertas-Tato} \\
	Computer Systems Department\\
	Universidad Polit{\'e}cnica de Madrid\\
	Madrid, Spain \\
	\texttt{javier.huertas.tato@upm.es} \\	
	\And
	\href{https://orcid.org/
0000-0002-5051-3475}{\includegraphics[scale=0.06]{orcid.pdf}\hspace{1mm}David Camacho} \\
	Computer Systems Department\\
	Universidad Polit{\'e}cnica de Madrid\\
	Madrid, Spain \\
	\texttt{david.camacho@upm.es} \\		
}

\renewcommand{\headeright}{}
\renewcommand{\undertitle}{}
\renewcommand{\shorttitle}{Exploring Dimensionality Reduction Techniques in Multilingual Transformers}

\hypersetup{
pdftitle={Exploring Dimensionality Reduction Techniques in Multilingual Transformers },
pdfsubject={q-bio.NC, q-bio.QM},
pdfauthor={Á. Huertas-García, et al.},
pdfkeywords={Dimensionality Reduction, Natural Language Processing, Semantic Textual Similarity, Multilingual Transformers, Visualization},
}

\begin{document}
\maketitle

\begin{abstract}
Both in scientific literature and in industry, semantic and context-aware Natural Language Processing-based solutions have been gaining importance in recent years. The possibilities and performance shown by these models when dealing with complex Human Language Understanding tasks is unquestionable, from conversational agents to the fight against disinformation in social networks. In addition, considerable attention is also being paid to developing multilingual models to tackle the language bottleneck. The growing need to provide more complex models implementing all these features has been accompanied by an increase in their size, without being conservative in the number of dimensions required. This paper aims to give a comprehensive account of the impact of a wide variety of dimensional reduction techniques on the performance of different state-of-the-art multilingual Siamese Transformers, including unsupervised dimensional reduction techniques such as linear and nonlinear feature extraction, feature selection, and manifold techniques. In order to evaluate the effects of these techniques, we considered the multilingual extended version of Semantic Textual Similarity Benchmark (mSTSb) and two different baseline approaches, one using the pre-trained version of several models and another using their fine-tuned STS version. The results evidence that it is possible to achieve an average reduction in the number of dimensions of $\SI{91.58 \pm 2.59}{\percent}$ and $\SI{54.65 \pm 32.20}{\percent}$, respectively. This work has also considered the consequences of dimensionality reduction for visualization purposes. The results of this study will significantly contribute to the understanding of how different tuning approaches affect performance on semantic-aware tasks and how dimensional reduction techniques deal with the high-dimensional embeddings computed for the STS task and their potential for other highly demanding NLP tasks.	
\end{abstract}


\section{Introduction}
\label{sec:introduction}

Natural Language Processing (NLP) includes various disciplines that provide a system with the ability to process and interpret natural language, just like humans use language as a communication and reasoning tool~\cite{OtterDanielW2021ASot}. Due to the recent increases in computational power, parallelization, the availability of large data sets, and recent advances in artificial intelligence, especially in the Machine Learning research field, NLP has been steadily proliferating and has garnered immense interest~\cite{OtterDanielW2021ASot,TayYi2020ETAS}. In the very recent past, Transformer-based architectures~\cite{vaswani2017attention} have become an indispensable staple in the NLP field. Transformer models are able to capture latent syntactic-semantic information and encode text's meaning as contextual vectors in a high-dimensional space referred as embeddings~\cite{TayYi2020ETAS, devlin2018bert}. In contrast to previous approaches, such as Statistical Natural Language Processing, the use of the attention mechanism provided by these architectures allows to take into consideration a plethora of characteristics involved in human language.

The tremendous power of Transformer-based models in Natural Language Understanding (NLU) and the new models that are continuously being proposed allow to dramatically improve the state-of-the-art results in varied NLP tasks, including question answering, sentence classification, and sentence pair regression like Semantic Textual Similarity (STS)~\cite{TayYi2020ETAS,devlin2018bert,reimers2019sentencebert, bertuit_2022}. The semantic evaluation performed in this task is one of the ``levels of language'' that determines the possible meanings of a sentence by focusing on the interactions among word-level~\cite{NLP-chowdhury}. It entails a high degree of complexity, given the large amount of different characteristics involved. In STS tasks, the systems need to compute how similar are two sentences considering all these features, returning a similarity score, usually ranging between 0 and 5~\cite{Cer_2017}. 

Regarding sentence-pair regression tasks,
to overcome the massive computational overhead caused by the quadratic dependence on the input size of attention mechanism in Transformers models~\cite{reimers2019sentencebert,humeau2020polyencoders}, the use of siamese architectures is a very effective method for deriving semantically meaningful sentence embeddings in an efficient way~\cite{reimers2019sentencebert,humeau2020polyencoders}. This approach is also called dual-encoder, bi-encoder or siamese architecures. As explained by Humeau et al.~\cite{humeau2020polyencoders}, the training of a siamese architecture  consists of two pre-trained Transformer-based models with tied weights that can be fine-tuned for a specific task like computing separately semantic embeddings for a pair of sentences and measure their similarity using the extensively used cosine similarity function~\cite{zhelezniak2019correlation}. 

Despite the practical features of cosine similarity, such as symmetry and spatial interpretation, this similarity metric has complexity $O(N)$: time and memory grow linearly with the number of dimensions of the vectors compared~\cite{SIDOROV2014}. Thus, dimensionality is a bottleneck for similarity computation and embedding storage. Moreover, an increasing number of studies using ensemble approaches based on the concatenation of embeddings can be found in the literature~\cite{AraqueOscar2017Edls,ensemble-fake,KhanAsifullah2020Asot}, aiming to improve the results in state-of-the-art tasks but accentuating this issue. Given that the application of dimensionality reduction techniques can mitigate this bottleneck, it requires further exploration.

Although in the history of NLP, the focus has mainly been on proposing architectures for English tasks. Nevertheless, interest in developing multilingual NLP tools has grown recently. The incorporation of tasks in various languages in the SemEval~\cite{semeval-2020-semantic} and CLEF~\cite{10.1007/978-3-030-28577-7_1} competitions are clear evidence of this, surmounting the language bottleneck, but also the number of new multilingual models constantly proposed~\cite{10.1007/978-3-030-28577-7_1}.  

The present paper seeks to address how different dimensionality reduction techniques impact the performance of pre-computed embeddings from multilingual siamese fashion Transformer-based models focusing on semantic similarity by employing four different approaches where the reduction techniques are compared with the pre-trained and fine-tuned versions of the models. 

The remainder of the article is organized as follows:
Section \ref{sec:related-work} outlines previous related work on dimensionality reduction techniques, their application in Deep Learning, and the importance of multilingual semantics in NLP. Section \ref{sec:methodology} describes the approaches followed in this research to evaluate the impact of dimensionality reduction techniques, the multilingual models and the dimensionality reduction techniques applied, and Section \ref{sec:exp-setup} the data and process used to fit and evaluate these techniques. The experimental results are discussed in Section \ref{sec:results}. Finally, Section 
\ref{sec:conclusion} summarizes the results of this work and draws conclusions. 

\section{Related Work}
\label{sec:related-work}

\subsection{Dimensionality Reduction Techniques}

Dimensionality reduction techniques aim to reduce the dimensionality of the data, removing irrelevant and redundant features while preserving critical information for subsequent applications, such as facilitating their visualization and understanding or leading to more compact models with better generalization ability~\cite{hyperspectral_2019, review_ft_select_2020}. 

There are different non-mutual exclusive criteria to classify dimensionality reduction techniques. 
Firstly, according to the reduction approach, these techniques can be classified into two categories: feature selection and feature extraction techniques~\cite{app_ft_ext_slect_2018}. Feature selection involves selecting a subset of original features that are useful for building models, effectively removing some of the less relevant features from consideration~\cite{PatelAnkurA2019Hulu, review_ft_select_2020}. On the other hand, feature extraction transforms the original data into another feature space with certain criteria, creating new variables by combining information from the original variables that capture much of the information contained in those original variables~\cite{app_ft_ext_slect_2018, hyperspectral_2019}. Additionally, feature extraction methods can be further subdivided into linear and nonlinear according to the variable combinations applied~\cite{HiraZenaM2015ARoF}. 

Secondly, according to the information available in the datasets, dimensionality reduction techniques can be classified as supervised and unsupervised~\cite{hyperspectral_2019,  review_ft_select_2020}. Supervised techniques require each data instance in the dataset to be labelled accordingly to the task, whereas unsupervised techniques are task-agnostic approaches and do not require labelled data.

\subsection{Dimensional reduction of embeddings}

In broad terms, embeddings can be categorised as pre-trained or downstream fine-tuned embeddings~\cite{pretrain_fine_tune_paradigma_2021}, and as  pre-computed or on-the-fly embeddings~\cite{on_the_fly_embeddings_2017,johnson2019billion}.

The first criteria to categorise embeddings is whether they come from pre-trained models for general tasks or are task-specific. Pre-trained embeddings are widely used as a starting point for downstream applications. A clear example is the current `Pre-train and Fine-tune' Paradigm of Transformer models~\cite{pretrain_fine_tune_paradigma_2021}. Training these models from scratch is prohibitively expensive in many cases. Alternatively, using self-supervised trained checkpoints of these models and their pre-trained embeddings as a starting point to be later fine-tuned for supervised downstream tasks is widely used. Unlike previous works in the literature that have only focused on reduced pre-trained embeddings~\cite{explore-embs-dim-red-2019,RaunakVikas2019ODLP,maria_mihaela_post-processing_2021}, in this work we are interested of the evaluation of the impact of dimensionality reduction on both types of embeddings, pre-trained and downstream fine-tuned embeddings.

Pre-computed embeddings in NLP are widespread, i.e., embeddings that may or may not be adjusted to a task but generated beforehand and not at each use time. A straightforward application of pre-computed embeddings is the semantic search NLP task for locating relevant information from massive amounts of text data~\cite{mitra_introduction_2018}. Semantic search task uses semantic textual similarity to compare an input text against a large set of texts from a database to extract relevant related information—usually, a distance metric like cosine similarity ranks which content should be extracted to an input query. However, computing the database embeddings each time a query is introduced is infeasible. Alternatively, it is preferable to compute the embeddings once, store them, and use these pre-computed embeddings for subsequent requests~\cite{johnson2019billion}. With this in mind, it is important to note the usefulness of reducing embeddings dimensions which can improve their utility in memory-constrained devices and benefit several real world applications.

As Camastra \& Vinciarelli mention~\cite{Camastra2008}, the use of more features than strictly necessary leads to several problems, pointing out that one of the main problems was the space needed to store the data. As the number of available information increases, the compression for storage purposes becomes even more critical. Additionally, for the scope of this work, it cannot be ignored that the application of dimensional reduction techniques for reducing pre-computed embeddings dimensions neither improves the runtime nor the memory requirement for running the models. It only diminishes the needed space to store embeddings and increases the speed to make computations (i.e. to calculate the cosine similarity between two vectors), which also contributes to decrease the considerable impact on the energy and carbon footprints generated during the production use of the models when pre-computed and stored embeddings are required~\cite{CO2_machine_learning_2019}. Research has tended to focus on implementing bigger and complex models rather than analysing methods to adjust the vector space to the desired task while being conservative on the number of dimensions required~\cite{CO2_machine_learning_2019}. An additional problem is that the storage of high-dimensional embeddings is challenging when dealing with large volume datasets~\cite{JDH17}. 

Recently, the study of Raunak et al.~\cite{explore-embs-dim-red-2019, RaunakVikas2019ODLP} has shed more light on the importance of reducing the size of embeddings produced by Machine Learning and Deep Learning models. More specifically, these authors draw attention to reducing the size of classical GloVe~\cite{pennington2014GloVe} and FastText~\cite{bojanowski2017fastext} pre-trained word embeddings using PCA-based post-processing algorithms, achieving similar or even better performance than the original embeddings. 

Other works on the potential of reducing pre-computed embeddings dimensions have been carried out~\cite{maria_mihaela_post-processing_2021} exploring the effect of Principal Components Analysis (PCA)~\cite{pca-original-1901, PCA_review} and Latent Semantic Analysis (LSA)~\cite{LSA_DeerwesterScott1990} dimensionality reduction techniques as a post-processing step of pre-trained GloVe word embeddings for text classification tasks. These authors also corroborated the usefulness of PCA for obtaining more accurate results with lower computational costs concluding that the PCA method is more suitable than LSA for dimensionality reduction. In the same way, Shimomoto et al.~\cite{ShimomotoEricaK2021Tcbo} propose solving topic classification and sentiment analysis by using PCA to transform pre-computed word embeddings of a text into a linear subspace. Their results showed the effectiveness of using the PCA subspace representation for text classification. This fact has already been proved by other authors~\cite{SongHaohao2020ECwR} showing that the storage, memory, and computation required by these large embeddings typically results in low efficiency in NLP tasks, pointing out the importance of other methods such as manifold learning to compress pre-computed and pre-trained GloVe and FastText embeddings.

Additionally, researchers have explored dimensional reduction techniques for visualizing semantic embeddings of small text corpora~\cite{visu-dim-red-2021}. The authors explored four dimension reduction strategies on pre-computed embeddings based on PCA and t-distributed stochastic neighbor embedding (t-SNE)~\cite{Hinton2002StochasticNE}, concluding that both methods preserved a significant amount of semantic information in the full embedding. Similarly, other study~\cite{de_rosa_how_2019} has focused on metaheuristic methods to find proper values for the perplexity t-SNE parameter and to optimize this word embeddings visualization. 

To the best of our knowledge, in the literature, dimension reduction research on embeddings has focused on using classical pre-computed word embeddings, including the popular GloVe or FastText embeddings. These classical word embeddings are static, word-level, and contextual-independent, and their main limitation is that they do not consider what context the word is being used. Moreover, the variety of embedding techniques explored is limited, focusing mainly on PCA. Likewise, these studies do not include multilingualism in their analyses, being limited to the English language. 

Hence, the presented work follows the research line proposed by different authors~\cite{explore-embs-dim-red-2019,RaunakVikas2019ODLP,maria_mihaela_post-processing_2021,visu-dim-red-2021} but takes a step forward, including a broader range of techniques and evaluating the capability of \textbf{dimensionality reduction techniques} in both \textbf{pre-trained and fine-tuned pre-computed embeddings} from state-of-the-art contextual-based Transformer models from the recently claimed \textbf{multilingual} point of view.

\subsection{Dimensionality Reduction Techniques in other fields}

The performance of Machine Learning (ML) models and, particularly, Deep Learning (DL) approaches is heavily dependent on the choice of data representation (or features) on which they are applied~\cite{dim_red_text_2013}. For that reason, much effort during deploying ML and DL solutions is dedicated to obtain a data representation that can support effective learning, where dimensionality reduction techniques can contribute.

In the literature, different examples of combining dimensionality reduction techniques with Deep Learning complex models can be found. CNNs models and PCA have already been combined for low-dimensional parameterization of complex 3D geomodels~\cite{LiuYimin20213CAd}. A hybrid architecture composed of PCA and Deep Neural Network models has also been proposed for time series forecasting to predict fine particulate matter concentrations in urban air pollution~\cite{ChoiSangWon2021APtd}. Additionally, dimensionality reduction techniques have also been applied as a pre-processing step to improve text document clustering~\cite{text_dim_red_app_2020}. Furthermore, 
recently the combination of Deep Learning models and dimension reduction techniques has also been explored in different specific application domains such as health science for cancer classification, where gene expression data is dimensionality reduced before training a Deep Belief Network (DBN) classifier~\cite{bio_dnn_dim_app_2021}. 

\subsection{Importance of Multilingual Semantics}
\label{sec:multi-semantic-apps}

Semantics has many applications in a wide range of domains and tasks. Recent developments regarding Information Retrieval tasks~\cite{mitra_introduction_2018, nogueira2020document, mSTSb-ref} have demonstrated the potential of combining semantic-aware models along with traditional baseline algorithms (e.g., BM25)~\cite{bm25}. Moreover, the use of semantic-aware models has proven to be an excellent approach to counteract informational disorders (i.e., misinformation, disinformation, malinformation, misleading information, or any other kind of information pollution)~\cite{wardle_information_2017,CarmiElinor2020DcRd, 9120902, huertas2021civic} or to build automated fact-checking approaches~\cite{martin2021factercheck}. Additionally, semantic similarity can be applied to organize data according to text properties, which is formally an unsupervised thematic analysis~\cite{grootendorst2020bertopic}. Following the same criterion, the measurement of semantic similarity between a sentence and each word in the sentence can be applied to extract the keywords with the highest semantic content from the sentence~\cite{grootendorst2020keybert}. All these applications rely on measuring semantic textual similarity (STS), making STS a crucial task in NLP. 

A key limitation of these semantic-aware solutions is the the language bottleneck~\cite{reimers_making_2020}. Language constitutes one of the most significant barriers to be addressed since a model's ability to handle multiple languages is essential for its widespread applications. 

Altogether, this work aims to broaden our knowledge of semantic-aware transformer-based models by analysing the impact of different dimensionality reduction techniques on the performance of multilingual siamese fashion Transformers on semantic textual similarity multilingual tasks. The results of this study will significantly contribute to understanding how different tuning approaches affect performance on semantic-aware tasks and how dimensional reduction techniques deal with the high dimensional embeddings computed for the STS task.

\section{Methodology}
\label{sec:methodology}

\begin{figure*}[t]
    \centering
    \includegraphics[width=0.6 \linewidth]{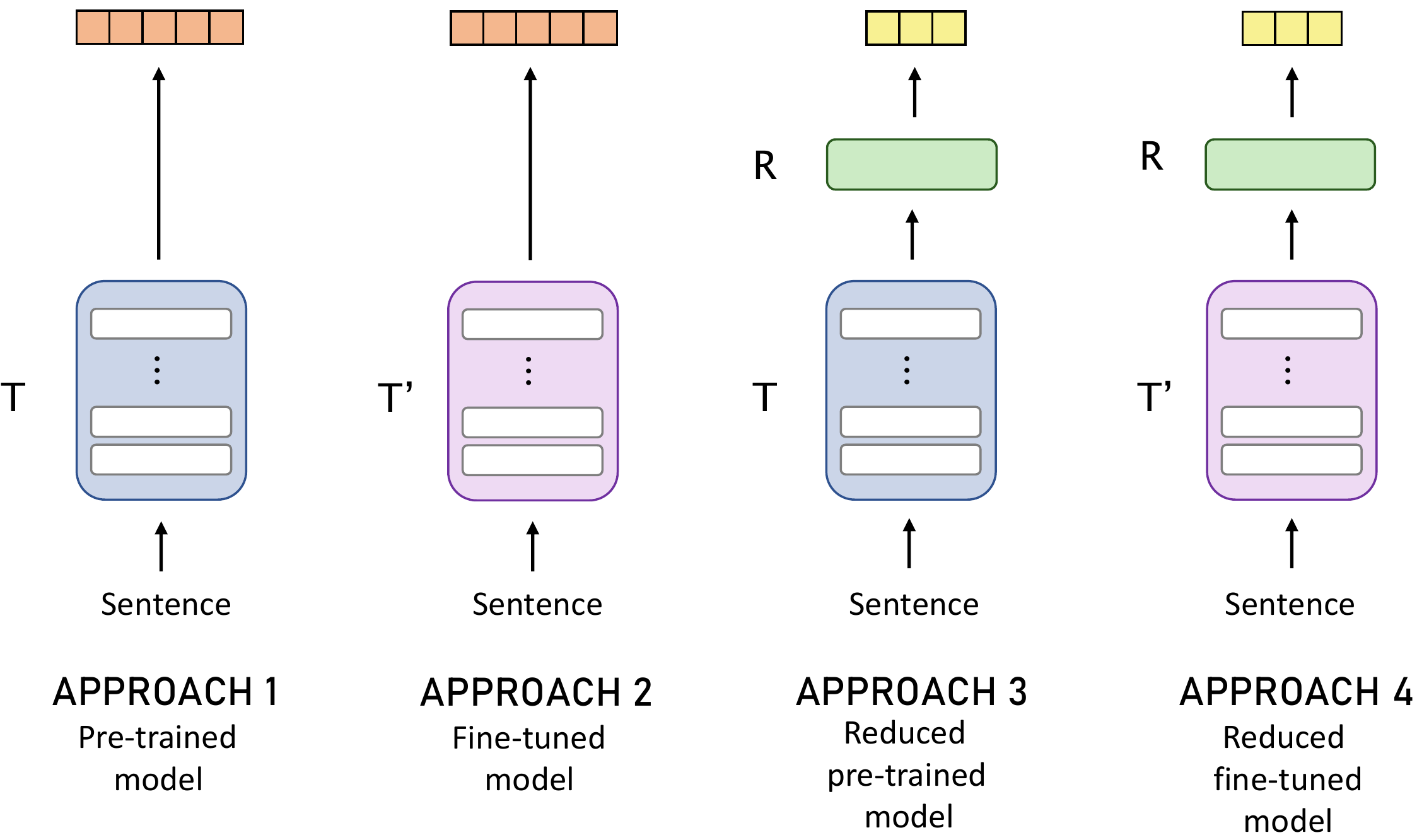}
    \caption{Representation of the approaches followed to evaluate the impact of different dimensionality reduction techniques in multilingual Transformers. Where \textit{T} represents the pre-trained Transformer model, \textit{T'} the fine-tuned Transformer model, and \textit{R} the dimensionality reduction technique.}
    \label{fig:Approaches_fig}
\end{figure*}

The main goal of this research lies in providing a deep analysis of the use of different dimensionality reduction techniques to reduce the size of the output embedding of different Transformer models and their impact in the performance. We have followed four approaches, aiming to evaluate and quantify the reduction margin and its effect in the performance while using different training methodologies. These four approaches are displayed in Figure~\ref{fig:Approaches_fig} and are described as follows:

\begin{itemize}
    \item{\textbf{Approach 1 - Pre-trained models}}. In the first approach we employ and directly evaluate the pre-trained models in the mSTSb task without applying any dimensionality reduction. This approach is used as baseline for general pre-trained models. 
    
    \item{\textbf{Approach 2 - Fine-tuned models}}. In this second approach, the pre-trained models are fine-tuned downstream and evaluated in the mSTSb task without applying any dimensionality reduction technique. This approach is used as baseline for fine-tuned models.
    
    \item{\textbf{Approach 3 - Reduced pre-trained models}}. In this approach, the embeddings generated by the pre-trained models from Approach 1 in the mSTSb train split are used to fit the different dimension reduction techniques and evaluate them in the mSTSb test split. Thus, an analysis between the results achieved in Approach 1 and Approach 3 will help to understand the impact of dimensionality reduction techniques in pre-trained models
    
    \item{\textbf{Approach 4 - Reduced fine-tuned models}}. This approach is equivalent to Approach 3 but uses the fine-tuned models in Approach 2, allowing to assess the impact of dimensionality reduction techniques in this type of models.
    
\end{itemize}

Next subsections describe in detail the techniques explored to reduce embeddings. Then, the multilingual transformer models used are presented.

\subsection{Dimensionality Reduction Techniques}


Transformer-based models embed textual information into vectors of high dimensionality. These pre-trained multilingual models usually generate embeddings with 768 dimensions, 1024 in case of  \textit{xlm-roberta-large}. 

As previously mentioned, dimensionality reduction techniques can be grouped according to two nonmutual exclusive criteria. In this project, we have included a whole range of types of dimensionality reduction techniques, including linear and nonlinear feature extraction and feature selection techniques. Nevertheless, since the Transformers models included in this work are employed in a siamese fashion to determine the degree of similarity between a pair of sentences, they output a pair of non-concatenative embeddings between which the similarity is estimated using the cosine distance. Hence, for each labelled similarity score, there are two non-concatenative embeddings. For this reason, even though we have labelled the data, only unsupervised methods are explored.

The dimensionality reduction techniques explored in this project are: 


\begin{itemize}
    \item{\textbf{Principal Component Analysis (PCA)}}: Principal Component Analysis \cite{pca-original-1901, PCA_review} is a powerful unsupervised linear feature extraction technique that computes a set of orthogonal directions from the covariance matrix that capture most of the variance in the data~\cite{muller_introduction_2001}. This is, it creates new uncorrelated variables that maximize variance, and at the same time, most existing structure in the data is retained.
    
    
    
    \item{\textbf{Independent Component Analysis (ICA)}}~\cite{hyvarinen_independent_2013}: Independent Component Analysis is an unsupervised feature extraction probabilistic method for learning a linear transformation to find components that are maximally independent between them and non-Gaussian (non-normal), but at the same time, they jointly maximize mutual information with the original feature space. 
    
    \item{\textbf{Kernel Principal Components Analysis (KPCA)}} \cite{scholkopf_nonlinear_1998}: Kernel-based learning method for PCA. It makes use of kernel functions to construct a nonlinear version of PCA linear algorithm by first implicitly mapping the data into a nonlinear feature space and then performing linear PCA on the mapped patterns~\cite{muller_introduction_2001}. The kernels considered in this project are the Polynomial, Gaussian RBF, Hyperbolic Tangent (Sigmoid), and Cosine kernels. 
    

    \item{\textbf{Variance Threshold}}: Unsupervised feature selection approach that removes all features with a variance below a threshold. Indeed, this technique selects a subset of features with large variances, considered more informative, without considering the desired outputs.

    \item{\textbf{Uniform Manifold Approximation and Projection for Dimension Reduction (UMAP)}}: The authors of UMAP~\cite{mcinnes2018umap-software} describe it is an algorithm that can be used for unsupervised dimension reduction based on manifold learning techniques and topological data analysis. In short, it first embeds data points in a new nonlinear fuzzy topological representation using neighbor graphs. Secondly, it learns a low-dimensional representation that preserves the maximum information of this space, minimizing Cross-Entropy. Compared to its counterparts, such as t-SNE, UMAP is fast, scalable, and allows better control of the desired balance between the local and global structure to be preserved. Two main parameters play a vital role in controlling this: (1) the number of sample points that defines a local neighborhood in the first step, and (2) the minimum distance between embedded points in low-dimensional space to be clustered in the second step. Larger values of the number of neighbors tend to preserve more global information in the manifold as UMAP has to consider larger neighborhoods to embed a point. Likewise, larger minimum distance values prevent UMAP from packing points together and preserve the overall topological structure.

\end{itemize}

\newcolumntype{L}{>{\RaggedRight\arraybackslash}X}
\newcolumntype{Y}{>{\centering\arraybackslash}X}

\begin{table}[htpb]
\centering
\caption{Considerations about the previous scaling steps and the characteristics of the different dimensionality reduction techniques applied in this project.}
\begin{tabular}{cccccc} 
\toprule
                   & \textbf{PCA}             & \textbf{KPCA}            & \textbf{ICA}  &
                  \begin{tabular}[c]{@{}c@{}}\textbf{Variance}\\\textbf{Threshold}\end{tabular} & \textbf{UMAP}           \\ 
\midrule 
Preprocessor       & 

\begin{tabular}[c]{@{}c@{}}Standard\\ \end{tabular}

& \begin{tabular}[c]{@{}c@{}}Standard\\ \end{tabular}

&  & 
\begin{tabular}[c]{@{}c@{}}MinMax \\ \end{tabular}
                                              & \begin{tabular}[c]{@{}c@{}}MinMax \\ \end{tabular}  \\
\hline  \noalign{\vskip 1ex}                                            
Scalation          & \xmark               & \xmark               & \xmark    & \xmark                                                           & \xmark              \\ [1ex]
\hline  \noalign{\vskip 1ex} 
Normalization      & \xmark               & \xmark               &      &                                                             &                \\ [1ex] 
\hline \noalign{\vskip 1ex}
Unsupervised       & \xmark               & \xmark               & \xmark    &   \xmark                                                          & \xmark              \\[1ex]
\hline \noalign{\vskip 1ex}
\begin{tabular}[c]{@{}c@{}}Feature \\ Selection \end{tabular}  & ~               &                 &      & \xmark                                                           &                \\[2ex] \hline \noalign{\vskip 1ex}
\begin{tabular}[c]{@{}c@{}}Feature \\ Extraction \end{tabular} & \xmark               & \xmark               & \xmark    &                                                             & \xmark              \\[2ex]
\hline \noalign{\vskip 1ex}
Linear             & \xmark               &                 & \xmark    &                                                             &                \\[1ex] \hline \noalign{\vskip 1ex}
Non Linear         &                 & \xmark               &      &                                                             & \xmark              \\
\bottomrule
\end{tabular}

\label{tb:consideration-dim-red}
\end{table}
\newcolumntype{L}{>{\RaggedRight\arraybackslash}X}
\newcolumntype{Y}{>{\centering\arraybackslash}X}

\begin{table}[htpb]
\centering
\caption{Parameters with non-default values used in the previous scaling steps and the dimensionality reduction techniques applied in this project.}

\def\arraystretch{1.25}%
\begin{tabular}{ll} 
\toprule
\textbf{Technique} & \textbf{Parameters  }                                                                                                                                             \\ 
\hline
ICA       & \begin{tabular}[c]{@{}l@{}}random\_state = 0\\max\_iter = 320\\whiten = True\\tol = 5e-4\end{tabular}                                                            \\ 
\hline
KPCA      & \begin{tabular}[c]{@{}l@{}}kernels = [sigmoid, polynomial, rbf, cosine]\\eigen\_solver = arpack\\copy\_X = False\\random\_state = 0\end{tabular}                 \\ 
\hline
\begin{tabular}[c]{@{}c@{}} Variance \\ Threshold\end{tabular} 
  & threshold = [Min, Max, Decile of variance]                                                                                                                             \\ 
\hline
UMAP      & \begin{tabular}[c]{@{}l@{}}precomputed\_knn = True\\metric = cosine\\min\_dist = 1\\n\_neighbors = [5, 10, 50, 100, 125]\\angular\_rp\_forest = True\end{tabular}  \\
\bottomrule
\end{tabular}

\label{table:dim_red_param}
\end{table}

According to the previous preprocessing steps required before dimensionality techniques, it should be noted that PCA and KPCA assume a Gaussian distribution, and the features must be normalized; otherwise, the variances will not be comparable. Therefore, the StandardScaler is applied beforehand. Regarding ICA, non-Gaussian distribution is assumed, and the data is already withened by the algorithm, so no previous preprocessing step is necessary. For the Variance Threshold, the best standardization method is MinMaxScaler, as it transforms all features to the same scale but does not alter the initial variability. This allows the variance selection threshold set to affect all dimensions equally. Finally, since there is no Gaussian assumption under UMAP and the cosine distance calculation benefits from scaling the features to a given range, MinMaxScaler is applied before UMAP. A summary about the necessary considerations about the above scaling steps and the characteristics of the dimensionality reduction techniques applied in this project are listed in Table~\ref{tb:consideration-dim-red}.

Finally, it is worth mentioning that these dimensionality reduction techniques and preprocessing algorithms come from \textit{scikit-learn} v1.0.2~\cite{scikit-learn}, except for UMAP that belongs to \textit{umap-learn} v0.5.2~\cite{mcinnes2018umap-software}. For the sake of reproducibility, the different parameters and values used in the experiments are presented in Table~\ref{table:dim_red_param}. Finally, the variance threshold filters for the Variance Threshold technique tested are extracted by previously calculating the variance of each feature (i.e., 768 variances for 768 embedding dimensions), extracting the deciles, and including the maximum and minimum of these variances.

\subsection{Multilingual Models}
\label{ref:multi-models}

The effects of dimensionality reduction and fine-tuning process were explored in the following pre-trained multilingual models extracted from Hugging Face~\cite{wolf-etal-2020-transformers}:

\begin{itemize}

  \item \textbf{bert-base-multilingual-cased}: BERT~\cite{devlin2018bert} transformer model pre-trained on a large corpus of 104 languages Wikipedia articles using the self-supervised masked language modeling (MLM) objective with $\sim$177M parameters. 
  \item \textbf{distilbert-base-multilingual-cased}: Distilled version of the previous model, being on average twice as fast as this model, totalizing $\sim$134M parameters~\cite{Sanh2019DistilBERTAD}.
  \item \textbf{xlm-roberta-base}: Base-sized XLM-RoBERTa~\cite{xlm-roberta-2019} model totalizing $\sim$125M parameters. XLM-RoBERTa is RoBERTa model~\cite{liu2019roberta}, a robusted version of BERT,  pre-trained on CommonCrawl data containing 100 languages.  
  \item \textbf{xlm-roberta-large}: Large-sized XLM-RoBERTa~\cite{xlm-roberta-2019} model totalizing $\sim$355M parameters.
  \item \textbf{LaBSE}: Language-agnostic BERT Sentence
Embedding~\cite{feng2020languageagnostic} model trained for encoding and reducing the cosine distance between translation pairs with a siamese architecture based on BERT, a task related to semantic similarity. It was trained over 6 Billion translation pairs for 109 languages. The authors also reported that it has zero-shot capabilities as was able to produce decent results for other not seen languages.

\end{itemize}

\section{Experimental Setup}
\label{sec:exp-setup}

\subsection{Data}

The multilingual extended STS Benchmark (mSTSb)~\cite{mSTSb-ref} train set is used for fine-tuning the multilingual Transformers and fitting the variety of dimensional redcution techniques. This split consists of $16$ languanges\footnote{ar, cs, de, en, es, fr, hi, it, ja, nl, pl, pt, ru, tr, zh-CN, zh-TW} combined in $31$ mono and cross-lingual tasks  with $5,479$ pair of sentences each one. Likewise, mSTSb test set is used to evaluate the performance of the  models obtained from the different approaches. The mSTSb test set is also composed of  $31$ multilingual tasks with $1,379$ pair of sentences per task. 

To evaluate the performance in mSTSb, the sentence embeddings for each pair of sentences are computed and the semantic similarity is measured using the cosine similarity metric. Then, the Spearman correlation coefficient ($\rho$ or $r_s$) is computed between the scores obtained and the gold standard scores, as it is recognised as an official metric used for semantic textual similarity tasks~\cite{reimers-etal-2016-task, wang-etal-2018-glue}. 

It is important to note that the mSTSb data variety available for fitting (i.e., train split) totals $+183$k sentences (i.e., $16$ languages with $5,749$ pair of sentences each one). For linear PCA, this dataset is too large to fit in memory. To manage this situation, an Incremental PCA approach~\cite{incremental_pca_2008} is applied, which simply fits the PCA in batches being independent of the number of input data samples, but still dependent on the input data features. 

Similarly, KPCA and UMAP are computationally more expensive than the linear counterparts~\cite{scholkopf_nonlinear_1998, muller_introduction_2001}. For this reason, these dimensionality reduction techniques were fitted using a subset of $10$k pair of sentences (i.e., $20k$ sentences), ensuring always the number of data instances is larger than the number of dimensions. In order to perform this subsampling, the following requirements were taken into account: (1) all $16$ languages must be equally represented, giving a total of $625$ sentence pairs for each language; (2) all sentences present in the original train split will be present at least once in some language; (3) the representation of the different sentence pairs must be as random as possible. Following these criteria, we perform a sampling based on assigning sentences to a randomly selected language until we reach the maximum number of sampled data. To avoid any bias in the order in which the sentences are assigned to the languages, the different sentence pairs are shuffled randomly at each iteration. As each language reaches the maximum data, that language is discarded. This ensures a random distribution of samples in each language but includes the full range of sentences present in the original train data.

\subsection{Dimensionality reduced model fitting and selection}

In summary, the process of applying dimensional reduction techniques (i.e., Approach 3 and 4) for each model has two steps. Firstly, each dimensionality reduction technique is fitted through the embeddings computed by the  multilingual models using mSTSb train split. Finally, mSTSb test set is used to evaluate the selected reduced number of features for each technique in each model. A wide range of number of dimensions are explored for each dimensionality reduction technique as shown in Section~\ref{sec:results}.

\subsection{Transformers fine-tuning}

The fine-tuning process of the models based on the siamese training strategy for Approach 2, was performed following the methodology described by Reimers et al.~\cite{reimers2019sentencebert, reimers_making_2020}. The following hyperparameters were optimized: number of epochs, scheduler, weight decay, batch size, warmup ratio and learning rate. The hyperparameters values explored and results from the experiments performed can be consulted in Weight and Biases~\footnote{\href{https://wandb.ai/huertas_97/Paper-Dim-Red/table?workspace=user-huertas_97}{https://wandb.ai/huertas\_97/Paper-Dim-Red}}.

\subsection{Statistical comparison}
Additionally, to test if the use of reduced embeddings has a significant impact on the performance in comparison to the baseline approaches, we compare the average Spearman correlation coefficient of the five multilingual siamese Transformer models (see Section \ref{ref:multi-models}) between each pair of baseline and reduced approaches (i.e., Approach 1 vs Approach 3, Approach 2 vs Approach 4). For this purpose, as we are comparing the same set of models in different approaches, the two-tailed paired T-test using a significance level of $0.05$ is conducted to test the null hypothesis of identical average Spearman correlation coefficient scores.

\section{Results}
\label{sec:results}

\newcolumntype{L}{>{\RaggedRight\arraybackslash}X}
\newcolumntype{Y}{>{\centering\arraybackslash}X}

\begin{table}[htpb]
\centering
\caption{Average Spearman $r_s$ correlation coefficient comparison between Approach 1 (Ap. 1) and best dimensional reduction technique in Approach 3 (Ap. 3) for the multilingual Transformers.}
\begin{tabular}{ccccc} 

\toprule
\textbf{Model    }                                                                    &\textbf{ Ap. 1 $r_s$} & \begin{tabular}[c]{@{}c@{}}\textbf{Best}\\\textbf{Technique}\end{tabular}  & \textbf{Dimensions} & \textbf{Ap. 3 $r_s$ } \\
\hline
\begin{tabular}[c]{@{}c@{}}bert-base-\\multilingual-cased\end{tabular}       & 0.4342              & ICA            & 209        & 0.5019               \\[3ex]
\begin{tabular}[c]{@{}c@{}}distilbert-base-\\multilingual-cased\end{tabular} & 0.4531              & ICA            & 169        & 0.523                \\[3ex]
xlm-roberta-base                                                             & 0.3274              & ICA            & 249        & 0.5269               \\[3ex]
xlm-roberta-large                                                            & 0.2855              & ICA            & 1024       & 0.5392               \\[3ex]
LaBSE                                                                        & 0.7096              & ICA            & 129        & 0.7488               \\
\bottomrule
\end{tabular}

\label{table:best-tech-ap3}
\end{table}

This section aims to summarise the effect of a wide variety of dimensionality reduction techniques on the performance of multilingual siamese Transformers by comparing the baseline approaches (i.e., Approaches 1 and 2) with the reduced approaches (i.e., Approaches 3 and 4) for each model independently. It must be noted that this work does not pretend to provide a comparative analysis between the different models presented in Section~\ref{ref:multi-models} or to identify the best model for this task. In contrast, this work is focused on applying these dimensionality reduction techniques for reducing the dimensionality of the models' embeddings. Thus, the application of different dimensionality reduction techniques does not affect the execution or the memory requirement for running the models. It only diminishes the needed space to store embeddings and increases the speed to compute the cosine similarity between them.


\begin{figure}[htpb]
     \centering
     \begin{subfigure}[b]{\columnwidth}
         \centering
         \includegraphics[width=0.54\columnwidth]{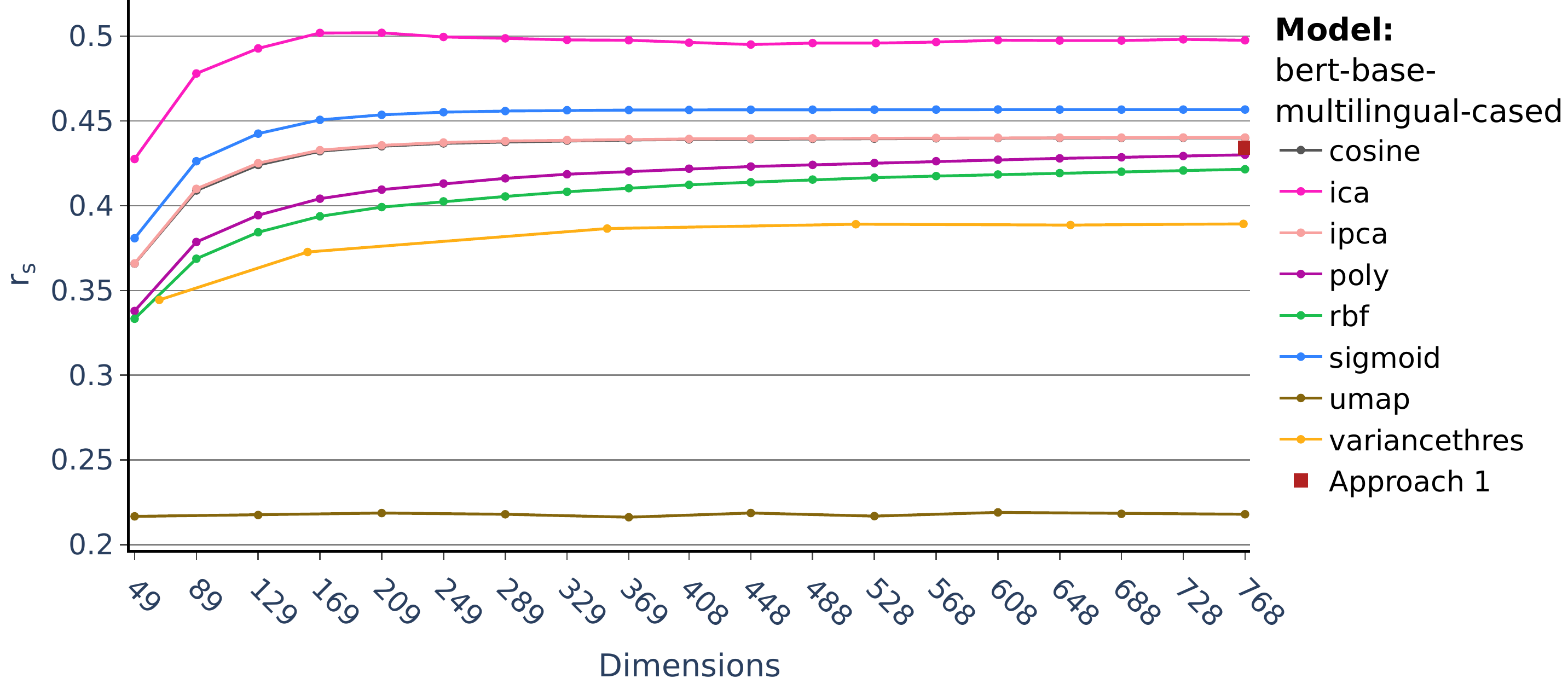}
         \caption{}
         \label{fig:dim-red-plot-ap3-bert}
     \end{subfigure}

     \begin{subfigure}[b]{\columnwidth}
         \centering
         \includegraphics[width=0.54\columnwidth]{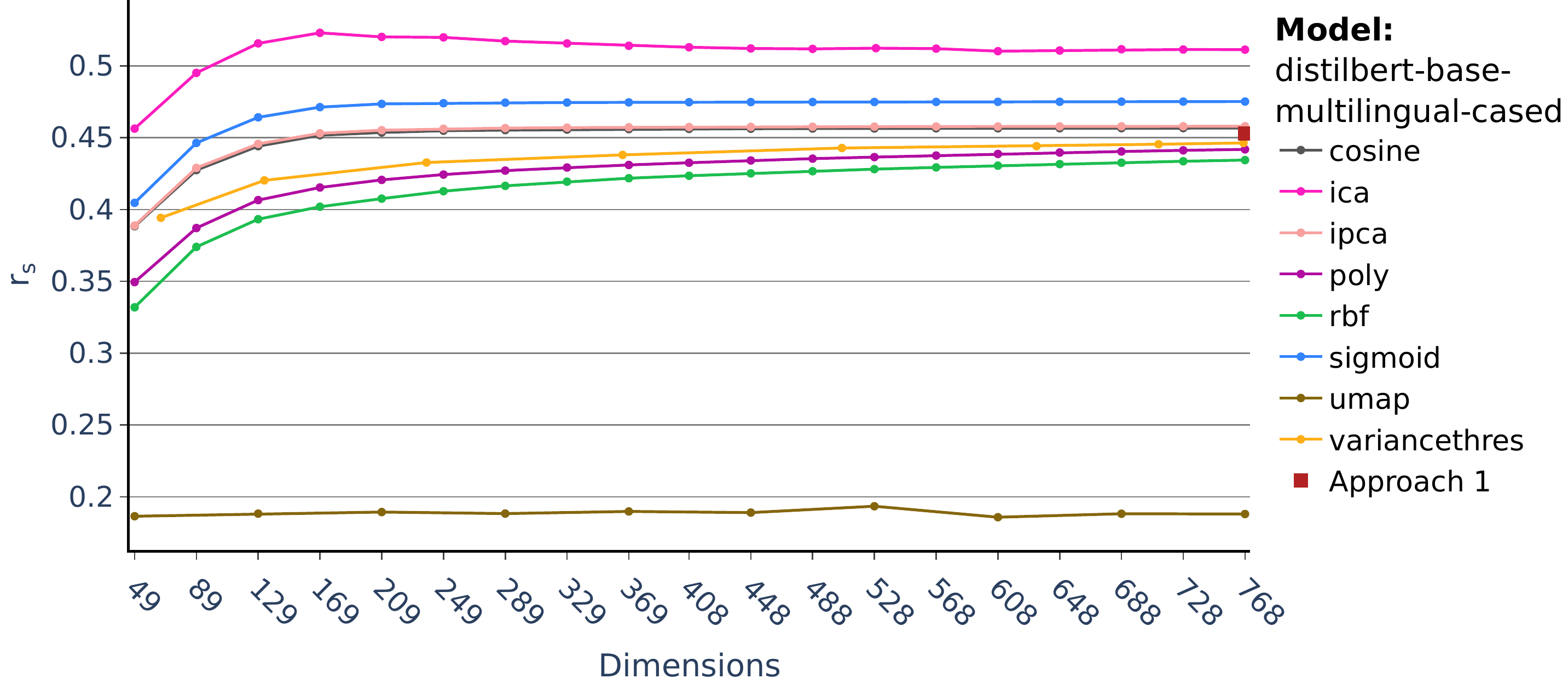}
         \caption{}
         \label{fig:dim-red-plot-ap3-distil}
     \end{subfigure}

     \begin{subfigure}[b]{\columnwidth}
         \centering
         \includegraphics[width=0.54\columnwidth]{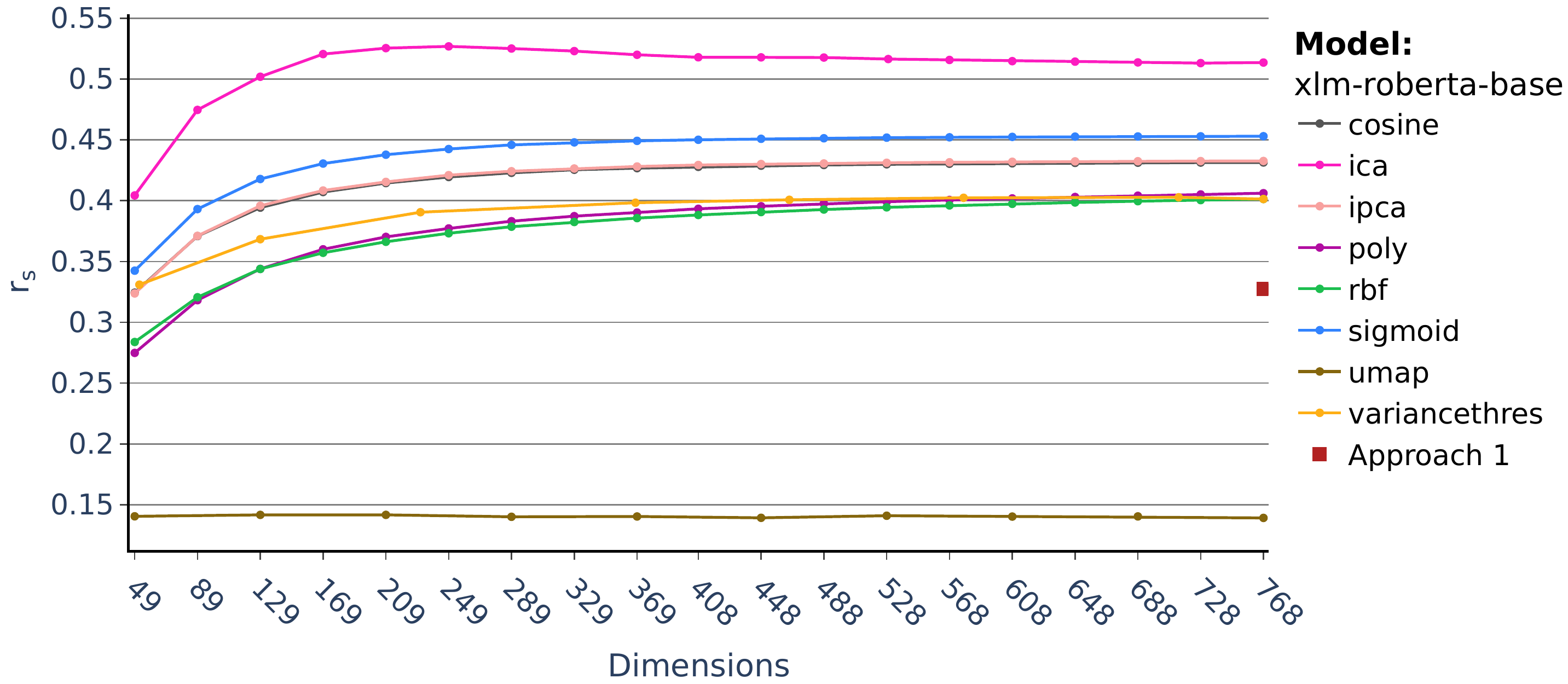}
         \caption{}
         \label{fig:dim-red-plot-ap3-xlm-base}
     \end{subfigure}
     
     \begin{subfigure}[b]{\columnwidth}
         \centering
         \includegraphics[width=0.54\columnwidth]{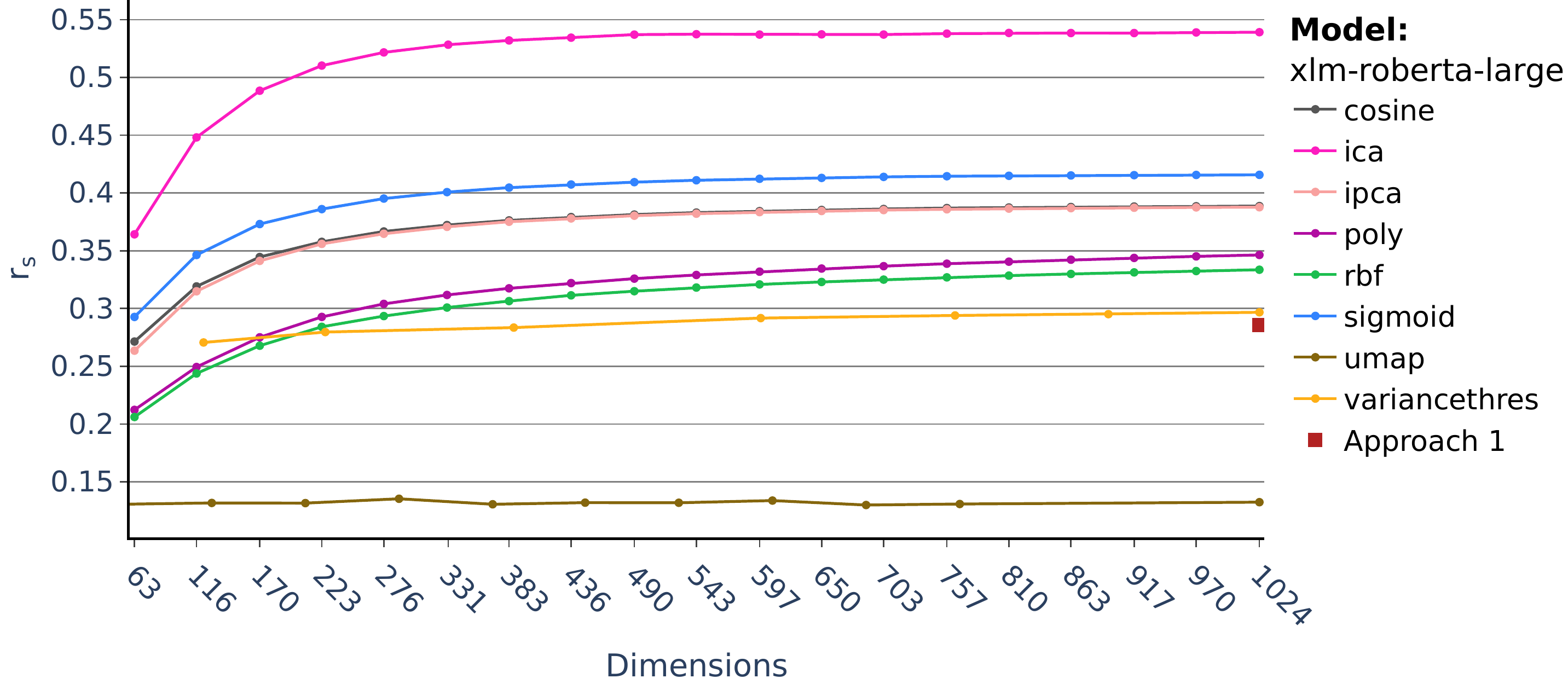}
         \caption{}
         \label{fig:dim-red-plot-ap3-xlm-large}
     \end{subfigure}

     \begin{subfigure}[b]{\columnwidth}
         \centering
         \includegraphics[width=0.54\columnwidth]{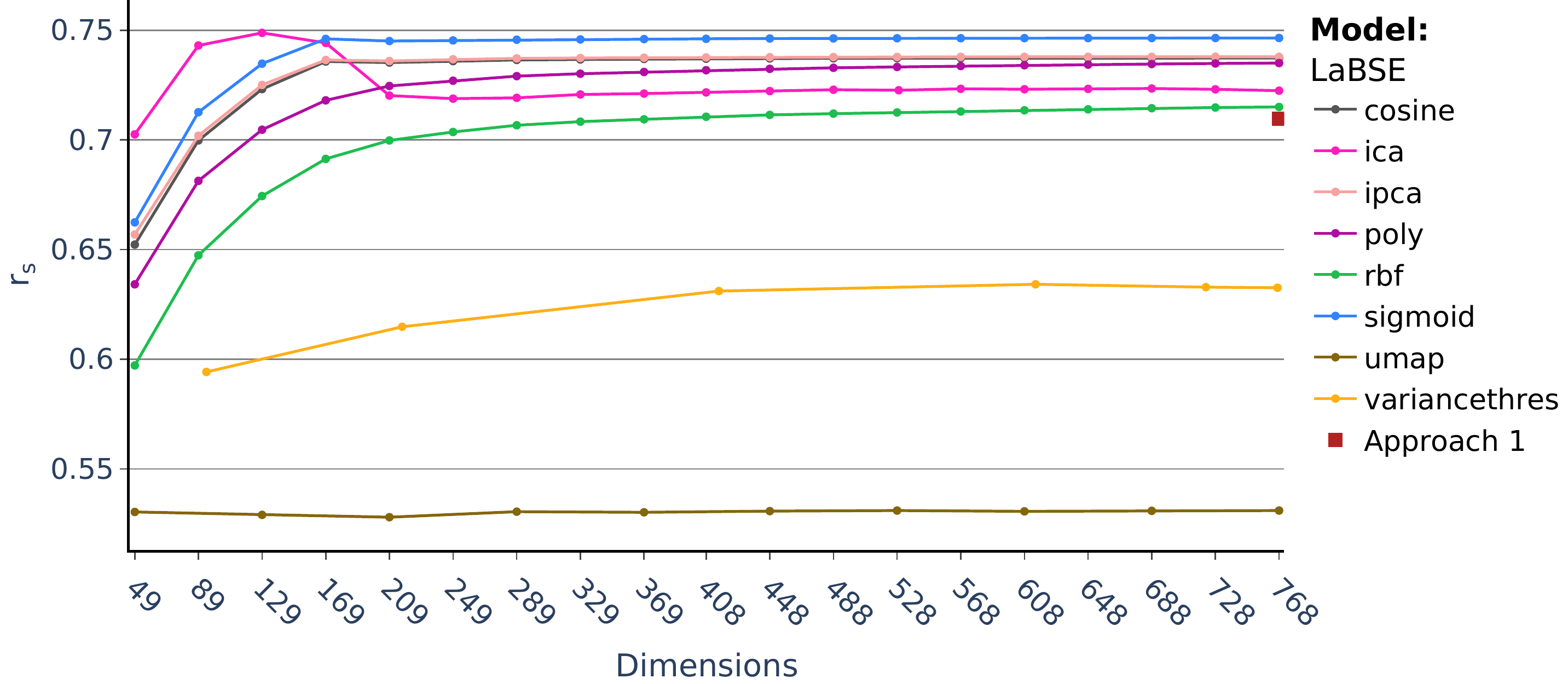}
         \caption{}
         \label{fig:dim-red-plot-ap3-labse}
     \end{subfigure}
     
\caption{Approach 3 average Spearman $r_s$ correlation coefficient in multilingual tasks from the mSTSb test as a function of the number of dimension for the different dimensionality reduction techniques.}
\label{fig:dim-red-plot-ap3}
\end{figure}

Due to space reasons, average results across the $31$ monolingual and cross-lingual tasks are presented instead of a breakdown by language. The average of Spearman correlation coefficients is computed by transforming each correlation coefficient to a Fisher's $z$ value, averaging them, and then back-transforming to a correlation coefficient.

\newcolumntype{L}{>{\RaggedRight\arraybackslash}X}
\newcolumntype{Y}{>{\centering\arraybackslash}X}

\begin{table}[htpb]
\centering
\caption{Analysis of the lowest reduced number of dimensions from Approach 3 that improves the result of the baseline Approach 1 for a specific performance threshold retained. For instance, 100\% threshold represents that the technique achieves at least the 100\% of the baseline approach score.}
\begin{tabular}{ccccc} 
\toprule
\begin{tabular}[c]{@{}c@{}}\textbf{Model}\\\textbf{(Ap. 1} \\ \textbf{Avg $r_s$)}\end{tabular}                            & \textbf{Technique} & \begin{tabular}[c]{@{}c@{}}\textbf{Threshold }\\\textbf{Performance}\\\textbf{Retained }\end{tabular} & \begin{tabular}[c]{@{}c@{}}\textbf{Dimensions~}\\\textbf{(\% reduction)}\end{tabular} & 
\begin{tabular}[c]{@{}c@{}}\textbf{Ap. 3~}\\\textbf{Avg $r_s$}\end{tabular}  \\ 
\hline
\multirow{8}{*}{\begin{tabular}[c]{@{}c@{}}bert-base-\\multilingual\\-cased \\(0.4342)\end{tabular}}       & IPCA               & 100\%                                                                                                 & 209 (73\%)                                                                            & 0.4251        \\
                                                                                                              & ICA                & \textbf{100\% }                                                                                                & \textbf{89 (88\%)}                                                                    & 0.4779        \\
                                                                                                              & poly               & 95\%                                                                                                  & 249 (68\%)                                                                            & 0.4130        \\
                                                                                                              & rbf                & 95\%                                                                                                  & 448 (42\%)                                                                            & 0.4138        \\
                                                                                                              & sigmoid            & 100\%                                                                                                 & 129 (83\%)                                                                            & 0.4425        \\
                                                                                                              & cosine             & 100\%                                                                                                 & 209 (73\%)                                                                            & 0.4350        \\
                                                                                                              & UMAP               & 50\%                                                                                                  & 129 (83\%)                                                                            & 0.2176        \\
                                                                                                              & VarThres           & 85\%                                                                                                  & 161(79\%)                                                                             & 0.3727        \\ 
\hline
\multirow{8}{*}{\begin{tabular}[c]{@{}c@{}}distilbert-base-\\multilingual\\-cased \\(0.4531)\end{tabular}} & IPCA               & 100\%                                                                                                 & 209 (73\%)                                                                            & 0.4553        \\
                                                                                                              & ICA                & \textbf{100\%}                                                                                                 & \textbf{49 (94\%)}                                                                    & 0.4564        \\
                                                                                                              & poly               & 95\%                                                                                                  & 369 (52\%)                                                                            & 0.4310        \\
                                                                                                              & rbf                & 95\%                                                                                                  & 608 (21\%)                                                                            & 0.4305        \\
                                                                                                              & sigmoid            & 100\%                                                                                                 & 129 (83\%)                                                                            & 0.4642        \\
                                                                                                              & cosine             & 100\%                                                                                                 & 209 (73\%)                                                                            & 0.4537        \\
                                                                                                              & UMAP               & 40\%                                                                                                  & 49 (94\%)                                                                             & 0.3942        \\
                                                                                                              & VarThres           & 95\%                                                                                                  & 238 (69\%)                                                                            & 0.438         \\ 
\hline
\multirow{8}{*}{\begin{tabular}[c]{@{}c@{}}xlm-roberta\\-base \\(0.3274)\end{tabular}}                     & IPCA               & 100\%                                                                                                 & 89 (88\%)                                                                             & 0.3711        \\
                                                                                                              & ICA                & \textbf{100\%  }                                                                                               & \textbf{49 (94\%)}                                                                    & 0.4043        \\
                                                                                                              & poly               & 100\%                                                                                                 & 129 (83\%)                                                                            & 0.3439        \\
                                                                                                              & rbf                & 100\%                                                                                                 & 129 (83\%)                                                                            & 0.3439        \\
                                                                                                              & sigmoid            & 100\%                                                                                                 & 49 (94\%)                                                                             & 0.3425        \\
                                                                                                              & cosine             & 100\%                                                                                                 & 89 (88\%)                                                                             & 0.3709        \\
                                                                                                              & UMAP               & 40\%                                                                                                  & 10 (99\%)                                                                             & 0.1320        \\
                                                                                                              & VarThres           & 100\%                                                                                                 & 52 (93\%)                                                                             & 0.3310        \\ 
\hline
\multirow{8}{*}{\begin{tabular}[c]{@{}c@{}}xlm-roberta\\-large\\(0.2885)\end{tabular}}                    & IPCA               & 100\%                                                                                                 & 116 (89\%)                                                                            & 0.3149        \\
                                                                                                              & ICA                & \textbf{100\%}                                                                                                 & \textbf{ 63 (94\%)}                                                                   & 0.3642        \\
                                                                                                              & poly               & 100\%                                                                                                 & 223~ (78\%)                                                                           & 0.2927        \\
                                                                                                              & rbf                & 100\%                                                                                                 & 276 (73\%)                                                                            & 0.2934        \\
                                                                                                              & sigmoid            & 100\%                                                                                                 & 63 (94\%)                                                                             & 0.2927        \\
                                                                                                              & cosine             & 100\%                                                                                                 & 116 (89\%)                                                                            & 0.3191        \\
                                                                                                              & UMAP               & 45\%                                                                                                  & 10 (99\%)                                                                             & 0.1365        \\
                                                                                                              & VarThres           & 100\%                                                                                                 & 598 (42\%)                                                                            & 0.2917        \\ 
\hline
\multirow{8}{*}{\begin{tabular}[c]{@{}cc@{}}LaBSE \\(0.7096)\end{tabular}}                                  & IPCA               & 100\%                                                                                                 & 129 (83\%)                                                                            & 0.7251        \\
                                                                                                              & ICA                & \textbf{100\%  }                                                                                               & \textbf{ 89 (88\%)}                                                                   & 0.7431        \\
                                                                                                              & poly               & 100\%                                                                                                 & 169 (78\%)                                                                            & 0.7181        \\
                                                                                                              & rbf                & 100\%                                                                                                 & 408 (47\%)                                                                            & 0.7106        \\
                                                                                                              & sigmoid            & 100\%                                                                                                 & 89 (88\%)                                                                             & 0.7127        \\
                                                                                                              & cosine             & 100\%                                                                                                 & 129 (83\%)                                                                            & 0.7232        \\
                                                                                                              & UMAP               & 70\%                                                                                                  & 10 (99\%)                                                                             & 0.5026        \\
                                                                                                              & VarThres           & 85\%                                                                                                  & 217 (72\%)                                                                            & 0.6148        \\
\bottomrule
\end{tabular}

\label{table:dim-red-fast-thres-ap3}
\end{table}

\newcolumntype{L}{>{\RaggedRight\arraybackslash}X}
\newcolumntype{Y}{>{\centering\arraybackslash}X}

\begin{table}[htpb]
\centering
\caption{Average Spearman $r_s$ correlation coefficient comparison between Approach 2 (Ap. 2) and best dimensional reduction technique in Approach 4 (Ap. 4) for the multilingual Transformers.}
\begin{tabular}{cccccc} 
\toprule
\textbf{Model}                                                                          & \textbf{Ap. 2 $r_s$} & \begin{tabular}[c]{@{}c@{}}\textbf{Best}\\\textbf{Technique}\end{tabular} & \textbf{Dimensions} & \textbf{Ap. 4 $r_s$}  
\\ 
\hline

\begin{tabular}[c]{@{}c@{}}bert-base-\\multilingual-cased-\\fine-tuned\end{tabular}       & 0.7045                       & ICA                     & 568                 & 0.7117                        
\\[5mm]

\begin{tabular}[c]{@{}c@{}}distilbert-base-\\multilingual-cased-\\fine-tuned\end{tabular} & 0.6863                       & VarThres                & 692                 & 0.6842                        
\\[5mm]

\begin{tabular}[c]{@{}c@{}}xlm-roberta-base-\\fine-tuned\end{tabular}                                                           & 0.7470                       & VarThres                & 673                 & 0.7495                        
\\[3mm]
 
\begin{tabular}[c]{@{}c@{}}xlm-roberta-large-\\fine-tuned\end{tabular}   & 0.8150                       & \begin{tabular}[c]{@{}c@{}}KPCA-\\sigmoid\end{tabular}         & 1024                & 0.8176                        
\\[3mm]

LaBSE-fine-tuned                                                                        & 0.8242                       & \begin{tabular}[c]{@{}c@{}}KPCA-\\sigmoid\end{tabular}             & 768                 & 0.8243                        \\
\bottomrule
\end{tabular}
\label{table:best-tech-ap4}
\end{table}

As can be seen in Figure \ref{fig:dim-red-plot-ap3}, for every model, the pre-trained performance on mSTSb (i.e., Approach 1) is improved using different dimensional reduction techniques. These results evidence that dimension reduction techniques are able to somehow adjust the knowledge present in the pre-trained embeddings to the semantic similarity task. This fact becomes even more significant in the case of LaBSE, a model with zero-shot capabilities trained on a task close to semantic similarity, which also greatly benefits from the use of dimension reduction techniques, increasing the Spearman correlation coefficient to 0.4 points (see Figure \ref{fig:dim-red-plot-ap3-labse} and Table \ref{table:best-tech-ap3}). For the rest of the models, it is equally remarkable that the dimension reduction techniques improve the pre-training performance, almost doubling score in the models with the XLM-RoBERTa architecture.

Clearly, the best technique in Approach 3 is ICA. Not only because it obtains the most remarkable improvement in pre-training performance for all models as shown in Table \ref{table:best-tech-ap3}, but also because it is the technique that most quickly and with the fewest dimensions overcomes the pre-trained models of Approach 1 (see Table \ref{table:dim-red-fast-thres-ap3}). From the results of this Table, the ICA technique improves the pre-trained Approach 1 performances reducing an average of $\SI{91.58 \pm 2.59}{\percent}$ of the initial dimensions retaining $\SI{100}{\percent}$ of the baseline Approach 1 performance. Remarkably, the two-tailed paired T-test comparing Approach 1 vs Approach 3 using the values for this technique from Table \ref{table:best-tech-ap3} resulted in $p = 0.041$, indicating that the improvement in performance is significant when using ICA as a dimensional reduction techniques for pre-trained embeddings. These findings corroborate the ideas of Raunak et al.~\cite{explore-embs-dim-red-2019}, who maintained that reduced word embeddings can achieve similar or better performance than original pre-trained embeddings.

The most likely explanation for these results is 
the difference in the objective of ICA from the other feature extraction techniques. Even though all of them transform the initial space through combinations of dimensions into a new space, ICA technique is based on the optimization of mutual information~\cite{bishop_2006, hyvarinen_independent_2013}. It tries to find a space where the new features (latent variables) are as independent as possible from each other but as dependent as possible from the initial space. Therefore, in the case of ICA, unlike other techniques such as PCA or KPCA, a higher number of components does not necessarily mean an increase in the information retained or an improvement in the result (as can be seen in Figures \ref{fig:dim-red-plot-ap3} and clearly in Figure \ref{fig:dim-red-plot-ap3-labse} where there is a decrease from 169 dimensions onwards). This would explain why a low number of dimensions would outperform Approach 1. 

Likewise, the fact that it is the technique that achieves the best results in Approach 3 in all models could be due to the assumptions and characteristics of both the ICA and the pre-trained embeddings. First, the pre-trained embeddings probably include non-relevant and noisy variables as these embeddings are not adjusted to the STS task. Secondly, since ICA is a technique in which the original variables are related linearly to the latent variables, but for which the latent distribution is non-Gaussian, the noise present in pre-trained embeddings agnostic of the STS task could be managed appropriately. 

Interestingly, these results also emphasize that the issue of non-Gaussianity is more relevant than the nonlinearity issue. Non-Gaussianity would be more important than how the initial variables are combined as the ICA technique outperforms both linear PCA and nonlinear KPCA. This is in good agreement with other studies comparing the performance of PCA and ICA as a method for feature extraction in visual object recognition tasks~\cite{ica_image_retrieva_2004, ica_face_recog_2005}. 

Additionally, the presence of noisy variables in the pre-trained embeddings would be also corroborated by the low scores obtained from the Variance Threshold feature selection technique, which completely depends on the original variables and cannot manage these noisy distributions. 

Consequently, ICA shows great properties to obtain compacted embeddings versions of pre-trained models with a large decrease of dimensions that improve the result in the task of semantic similarity at a multilingual level.

For all these reasons, we can understand unsupervised dimensionality reduction, specially ICA, as a method of fitting pre-trained models for downstream tasks. As it will be seen in the next section and as might be expected, this unsupervised dimensionality reduction downstream fitting is not comparable to a supervised fitting such as the fine-tuning of Approach 2. However, downstream fitting by unsupervised dimensionality reduction techniques may present interesting advantages such as the fact that being unsupervised is task agnostic resulting in models with higher generalizability and with a lower number of dimensions. Also, these dimensionality reduction techniques do not require GPUs and apply a more interpretable methodology than a Deep Learning model fine-tuning such as Transformers.

\newcolumntype{L}{>{\RaggedRight\arraybackslash}X}
\newcolumntype{Y}{>{\centering\arraybackslash}X}

\begin{table}[htpb]
\centering
\caption{Analysis of the lowest reduced number of dimensions from Approach 4 that improves the result of the baseline Approach 2 for a specific performance threshold retained. For instance, 100\% threshold represents that the technique achieves at least the 100\% of the baseline approach score.}
\begin{tabular}{cccccc} 
\toprule
\begin{tabular}[c]{@{}c@{}}\textbf{Model}\\\textbf{(Ap. 2}\\ \textbf{Avg $r_s$)}\end{tabular}                            & \textbf{Technique} & \begin{tabular}[c]{@{}c@{}}\textbf{Threshold }\\\textbf{Performance}\\\textbf{Retained }\end{tabular} & \begin{tabular}[c]{@{}c@{}}\textbf{Dimensions~}\\\textbf{(\% reduction)}\end{tabular} & \begin{tabular}[c]{@{}c@{}}\textbf{Ap. 4}\\\textbf{Avg $r_s$}\end{tabular}  \\ 
\hline
\multirow{8}{*}{\begin{tabular}[c]{@{}c@{}}bert-base-\\multilingual\\-cased\\-fine-tuned\\(0.7045)\end{tabular}}        & IPCA               & 95\%                                                                                                  & 49 (94\%)                                                                             & 0.6710                                                                      \\
                                                                                                                           & ICA                & \textbf{100\% }                                                                                       & \textbf{169 (78\%)}                                                                   & 0.7047                                                                      \\
                                                                                                                           & poly               & 95\%                                                                                                  & 129 (83\%)                                                                            & 0.6716                                                                      \\
                                                                                                                           & rbf                & 95\%                                                                                                  & 169 (78\%)                                                                            & 0.6738                                                                      \\
                                                                                                                           & sigmoid            & 100\%                                                                                                 & 329 (57\%)                                                                            & 0.7048                                                                      \\
                                                                                                                           & cosine             & 95\%                                                                                                  & 49 (94\%)                                                                             & 0.6707                                                                      \\
                                                                                                                           & UMAP               & 70\%                                                                                                  & 10 (99\%)                                                                             & 0.5398                                                                      \\
                                                                                                                           & VarThres           & 100\%                                                                                                 & 393 (53\%)                                                                            & 0.7046                                                                      \\ 
\hline
\multirow{8}{*}{\begin{tabular}[c]{@{}c@{}}distilbert-base-\\multilingual\\-cased \\-fine-tuned\\(0.6863)\end{tabular}} & IPCA               & \textbf{95\%}                                                                                         & \textbf{49 (94\%)}                                                                    & 0.6533                                                                      \\
                                                                                                                           & ICA                & \textbf{95\%}                                                                                         & \textbf{49 (94\%)}                                                                    & 0.6556                                                                      \\
                                                                                                                           & poly               & 95\%                                                                                                  & 129 (83\%)                                                                            & 0.6542                                                                      \\
                                                                                                                           & rbf                & 95\%                                                                                                  & 129 (83\%)                                                                            & 0.6520                                                                      \\
                                                                                                                           & sigmoid            & \textbf{95\%}                                                                                         & \textbf{49 (94\%)}                                                                    & 0.6601                                                                      \\
                                                                                                                           & cosine             & 95\%                                                                                                  & 89 (88\%)                                                                             & 0.6631                                                                      \\
                                                                                                                           & UMAP               & 75\%                                                                                                  & 10 (99\%)                                                                             & 0.5189                                                                      \\
                                                                                                                           & VarThres           & 95\%                                                                                                  & 66 (91\%)                                                                             & 0.6620                                                                      \\ 
\hline
\multirow{8}{*}{\begin{tabular}[c]{@{}c@{}}xlm-roberta\\-base\\-fine-tuned~\\(0.7470)\end{tabular}}                     & IPCA               & 95\%                                                                                                  & 49 (94\%)                                                                             & 0.7198                                                                      \\
                                                                                                                           & ICA                & 95\%                                                                                                  & 49 (94\%)                                                                             & 0.7208                                                                      \\
                                                                                                                           & poly               & 95\%                                                                                                  & 89 (88\%)                                                                             & 0.7112                                                                      \\
                                                                                                                           & rbf                & 95\%                                                                                                  & 129 (83\%)                                                                            & 0.7134                                                                      \\
                                                                                                                           & sigmoid            & \textbf{100\%}                                                                                        & \textbf{289 (62\%)}                                                                   & 0.7472                                                                      \\
                                                                                                                           & cosine             & 95\%                                                                                                  & 49 (94\%)                                                                             & 0.7195                                                                      \\
                                                                                                                           & UMAP               & 75\%                                                                                                  & 10 (99\%)                                                                             & 0.5724                                                                      \\
                                                                                                                           & VarThres           & 100\%                                                                                                 & 411 (46\%)                                                                            & 0.7491                                                                      \\ 
\hline
\multirow{8}{*}{\begin{tabular}[c]{@{}c@{}}xlm-roberta\\-large \\-fine-tuned\\(0.8150)\end{tabular}}                    & IPCA               & 95\%                                                                                                  & 63 (94\%)                                                                             & 0.7910                                                                      \\
                                                                                                                           & ICA                & 95\%                                                                                                  & 63 (94\%)                                                                             & 0.7950                                                                      \\
                                                                                                                           & poly               & 95\%                                                                                                  & 63 (94\%)                                                                             & 0.7774                                                                      \\
                                                                                                                           & rbf                & 95\%                                                                                                  & 63 (94\%)                                                                             & 0.7760                                                                      \\
                                                                                                                           & sigmoid            & \textbf{100\%}                                                                                        & \textbf{223 (78\%)}                                                                   & 0.8151                                                                      \\
                                                                                                                           & cosine             & 95\%                                                                                                  & 63 (94\%)                                                                             & 0.7916                                                                      \\
                                                                                                                           & UMAP               & 80\%                                                                                                  & 10 (99\%)                                                                             & 0.6584                                                                      \\
                                                                                                                           & VarThres           & 95\%                                                                                                  & 95 (91\%)                                                                             & 0.7936                                                                      \\ 
\hline
\multirow{8}{*}{\begin{tabular}[c]{@{}c@{}}LaBSE \\-fine-tuned\\(0.8242)\end{tabular}}                                  & IPCA               & 95\%                                                                                                  & 89 (88\%)                                                                             & 0.8014                                                                      \\
                                                                                                                           & ICA                & 95\%                                                                                                  & 89 (88\%)                                                                             & 0.7986                                                                      \\
                                                                                                                           & poly               & 95\%                                                                                                  & 89 (88\%)                                                                             & 0.7898                                                                      \\
                                                                                                                           & rbf                & 95\%                                                                                                  & 129 (83\%)                                                                            & 0.7932                                                                      \\
                                                                                                                           & sigmoid            & \textbf{100\%}                                                                                        & \textbf{728 (5\%)}                                                                    & 0.8243                                                                      \\
                                                                                                                           & cosine             & 95\%                                                                                                  & 89 (88\%)                                                                             & 0.8001                                                                      \\
                                                                                                                           & UMAP               & 80\%                                                                                                  & 23 (97\%)                                                                             & 0.6640                                                                      \\
                                                                                                                           & VarThres           & 95\%                                                                                                  & 227 (70\%)                                                                            & 0.7964                                                                      \\
\bottomrule
\end{tabular}
\label{table:dim-red-fast-thres-ap4}
\end{table}

\begin{figure}[htpb]
     \centering
     \begin{subfigure}[b]{\columnwidth}
         \centering
         \includegraphics[width=0.65\columnwidth]{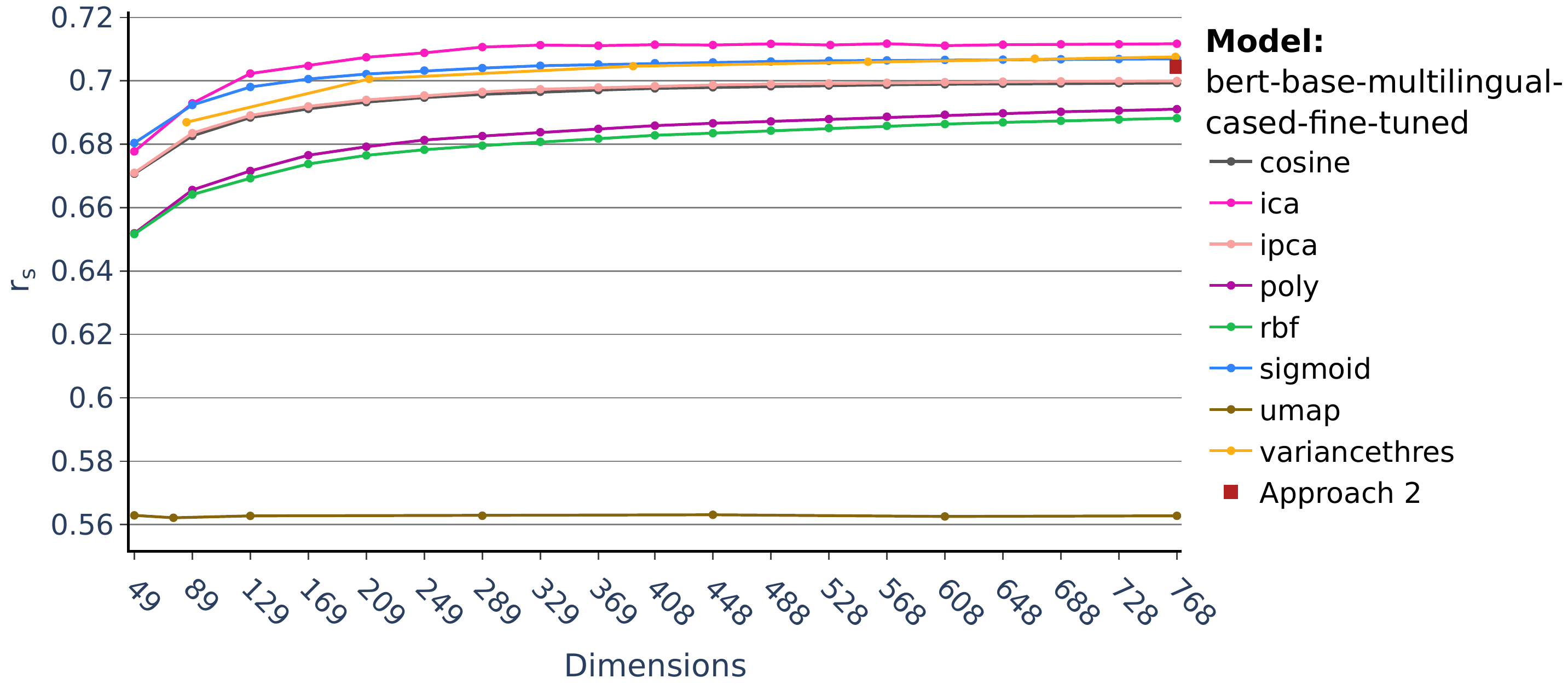}
         \caption{}
         \label{fig:dim-red-plot-ap4-bert}
     \end{subfigure}%
     \begin{subfigure}[b]{\columnwidth}
         \centering
         \includegraphics[width=0.65\columnwidth]{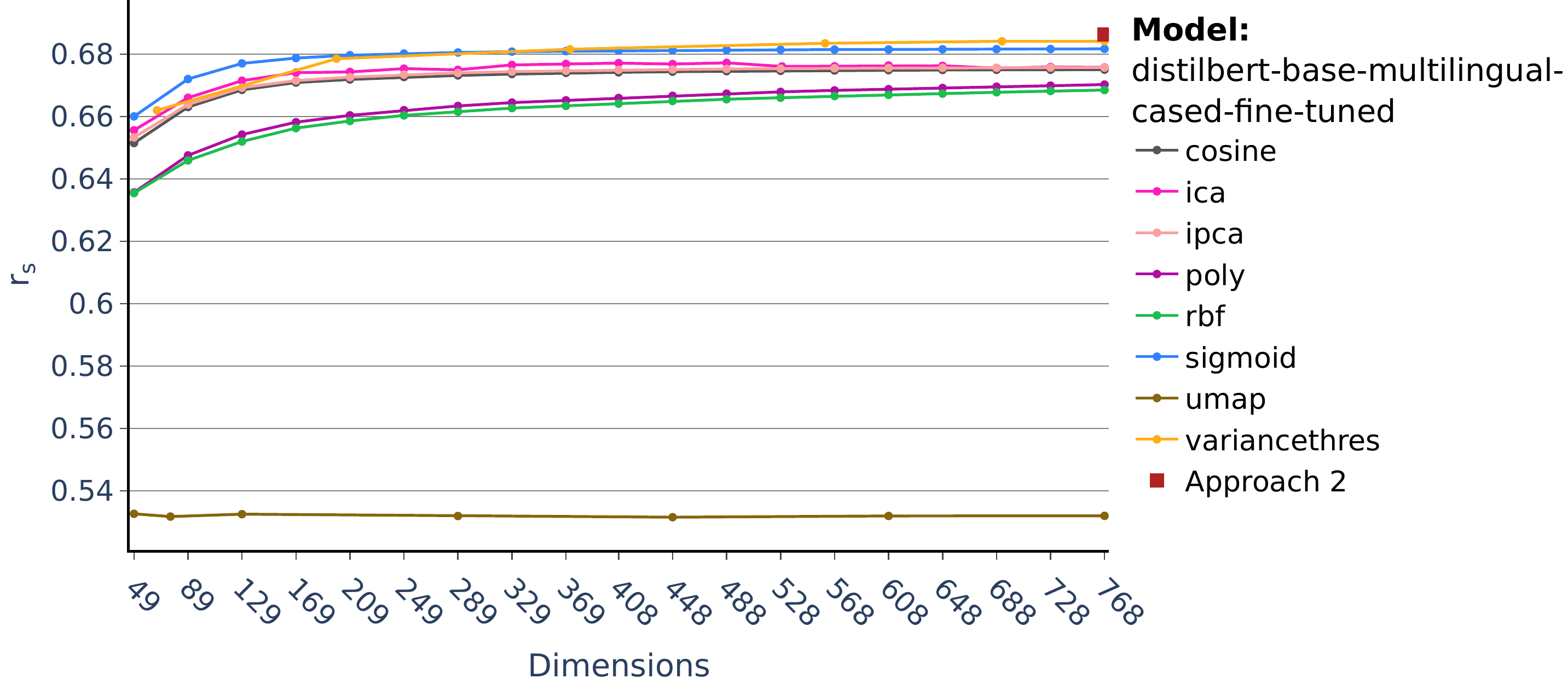}
         \caption{}
         \label{fig:dim-red-plot-ap4-distil}
     \end{subfigure}

     \begin{subfigure}[b]{\columnwidth}
         \centering
         \includegraphics[width=0.65\columnwidth]{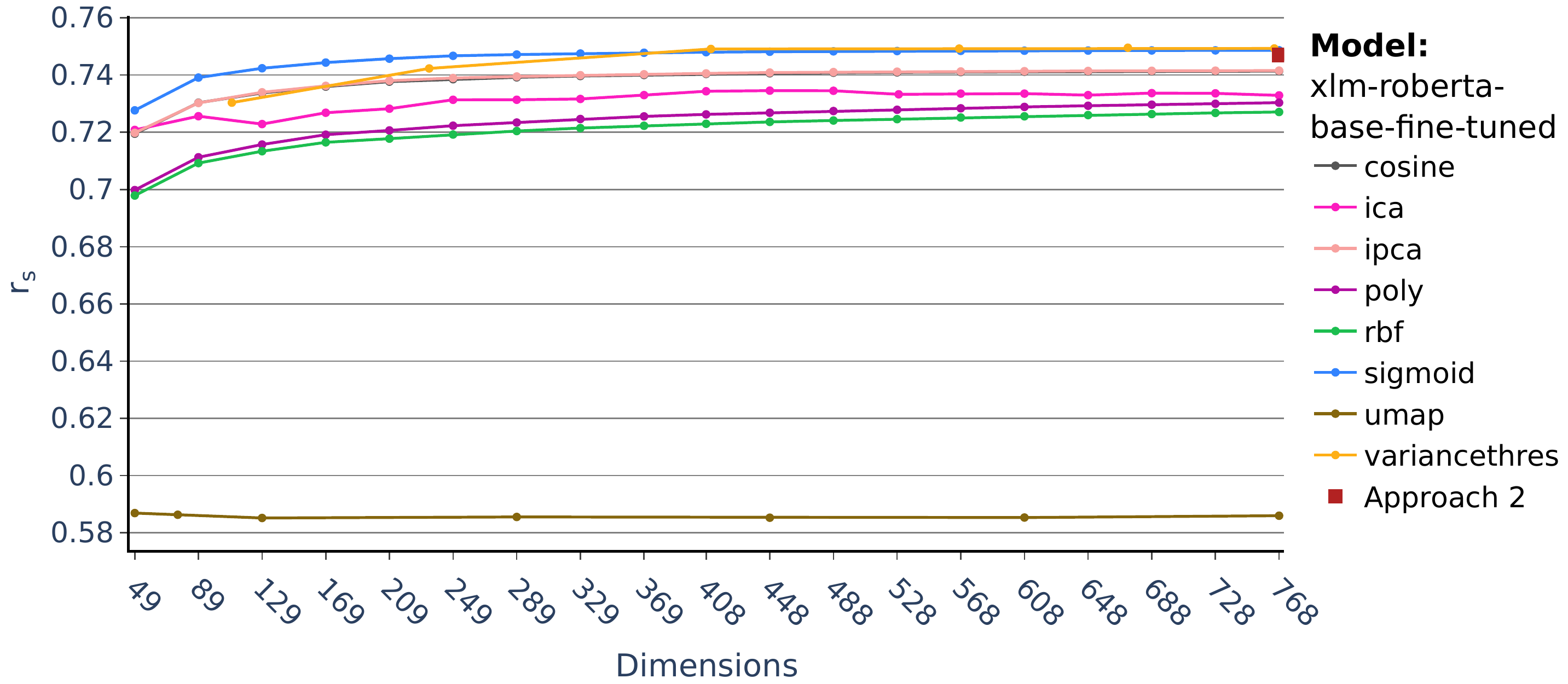}
         \caption{}
         \label{fig:dim-red-plot-ap4-xlm-base}
     \end{subfigure}
     
     \begin{subfigure}[b]{\columnwidth}
         \centering
         \includegraphics[width=0.65\columnwidth]{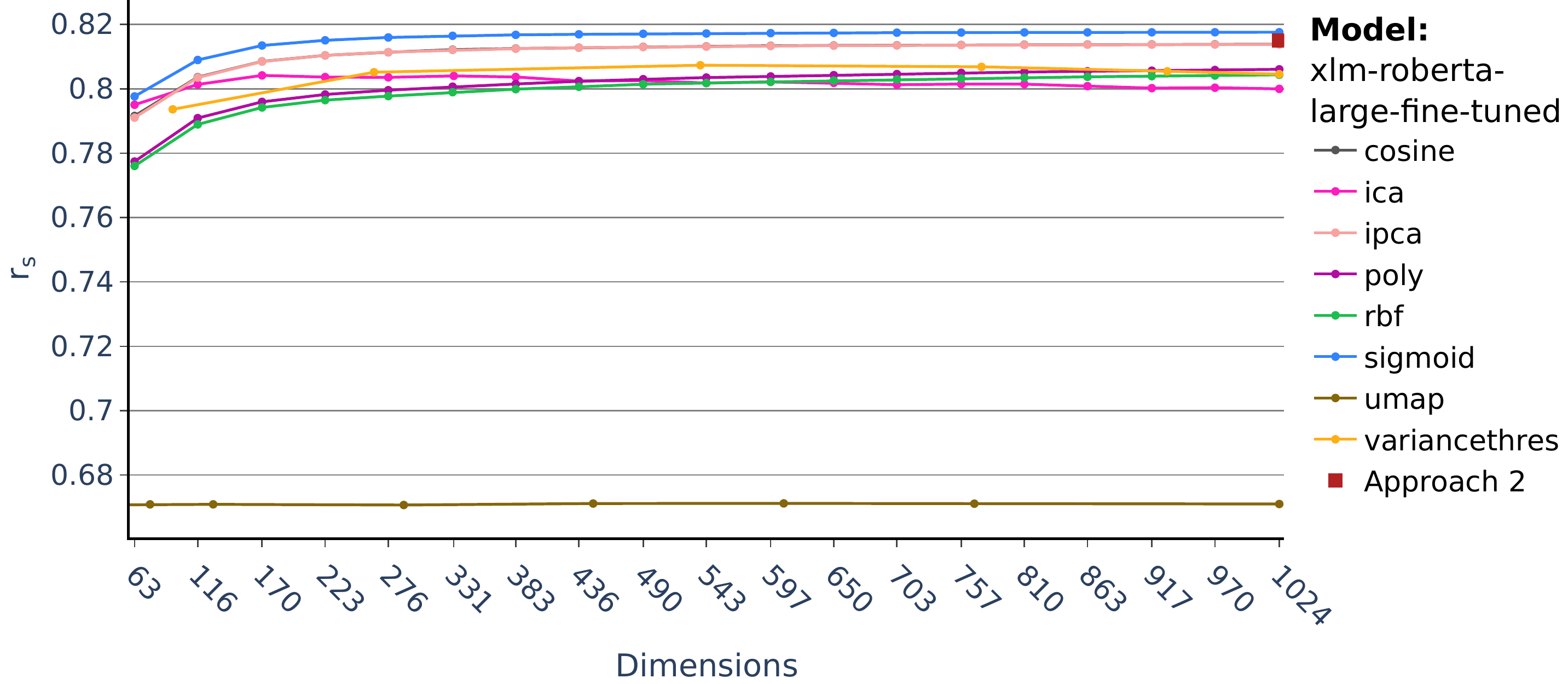}
         \caption{}
         \label{fig:dim-red-plot-ap4-xlm-large}
     \end{subfigure}

     \begin{subfigure}[b]{\columnwidth}
         \centering
         \includegraphics[width=0.65\columnwidth]{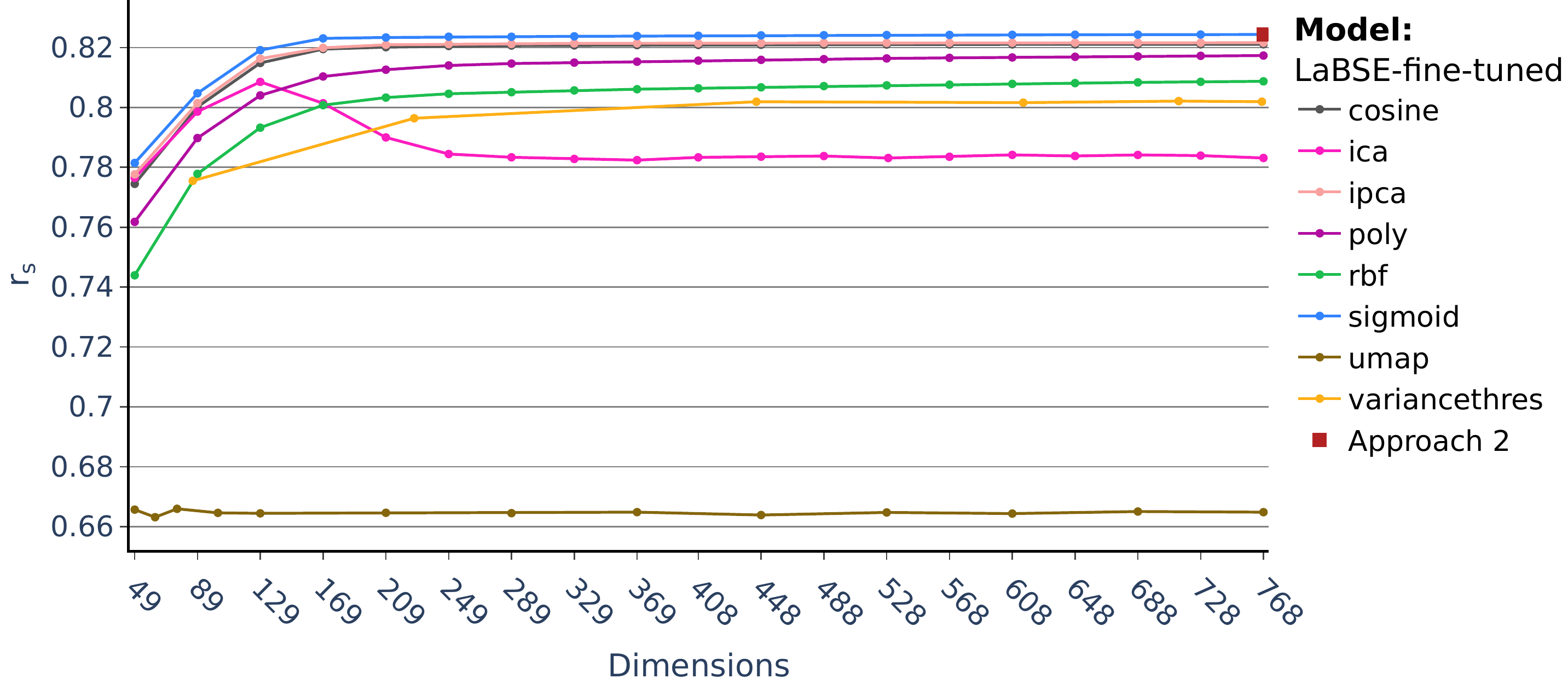}
         \caption{}
         \label{fig:dim-red-plot-ap4-labse}
     \end{subfigure}
     
\caption{Approach 4 average Spearman $r_s$ correlation coefficient in multilingual tasks from the mSTSb test as a function of the number of dimensions for the different dimensionality reduction techniques.}
\label{fig:dim-red-plot-ap4}
\end{figure}

\begin{figure}[t]
     \centering
     \begin{subfigure}[t]{.5\textwidth}
         \centering
         \includegraphics[width=0.9\linewidth]{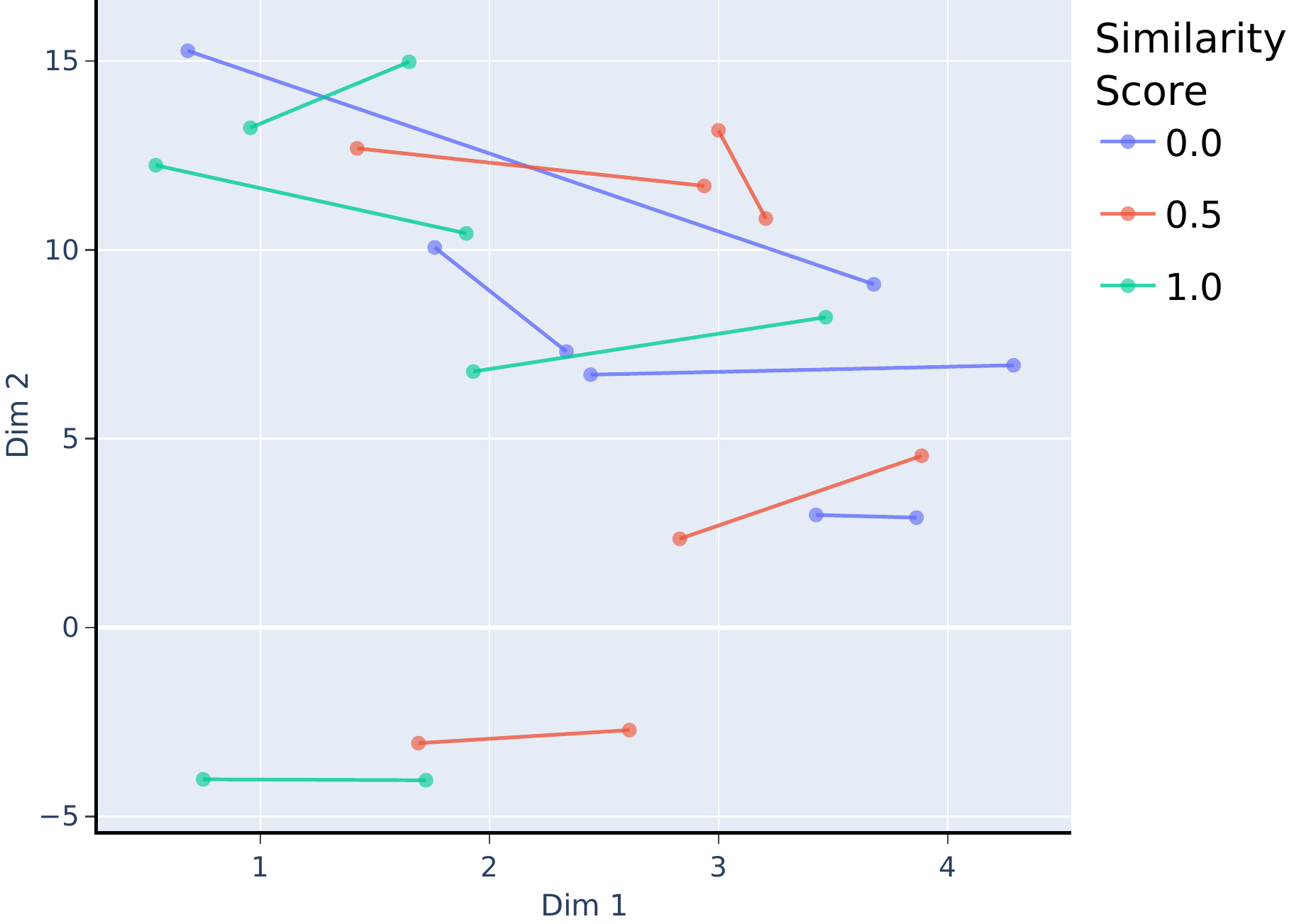}
         \caption{}
         \label{fig:bert-base-visu-example-pretrain}
     \end{subfigure}%
     \begin{subfigure}[t]{.5\textwidth}
         \centering
         \includegraphics[width=0.9\linewidth]{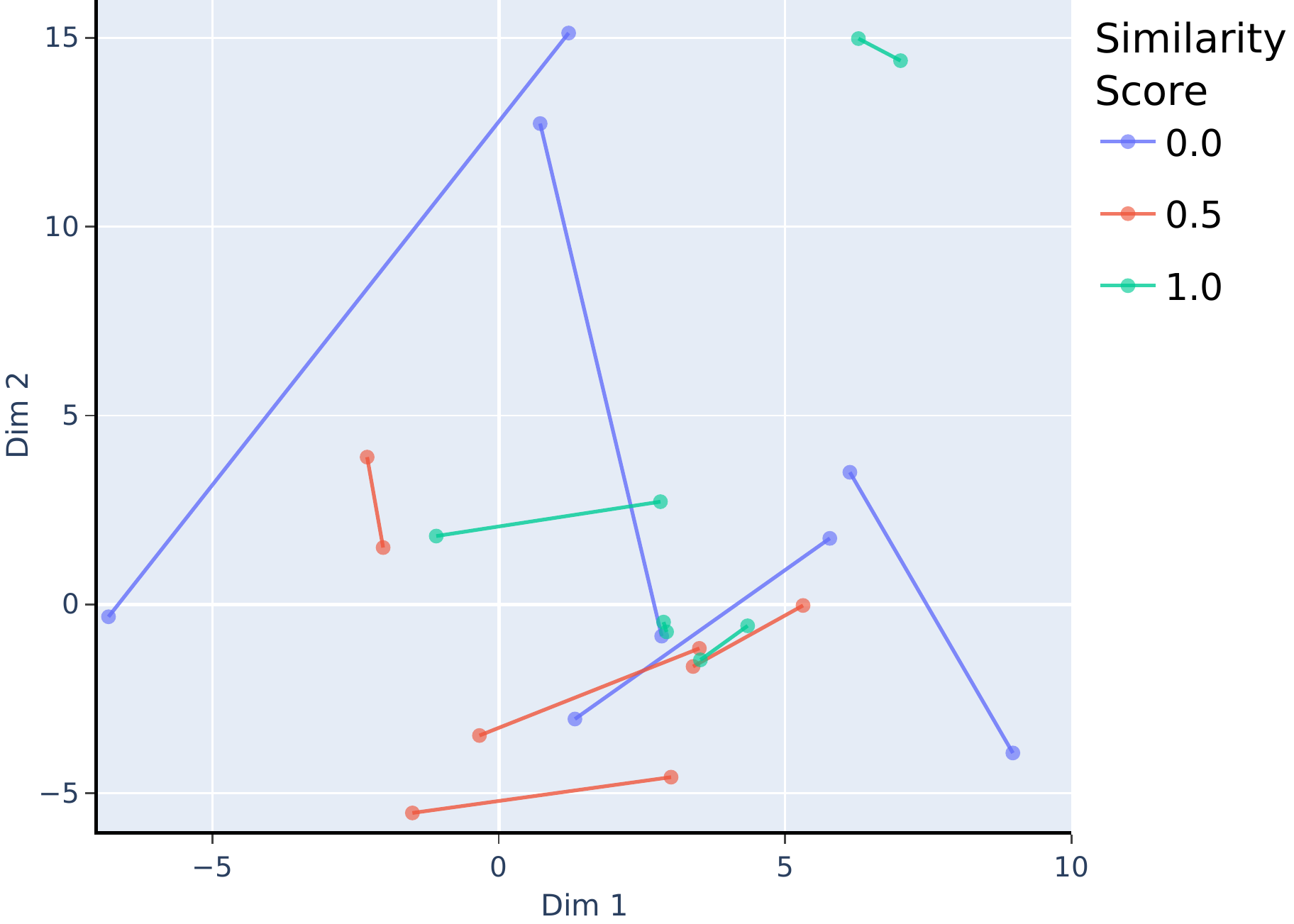}
         \caption{}
         \label{fig:bert-base-visu-example-fine-tuned}
     \end{subfigure}

\caption{Comparison of the representation of embeddings generated by the IPCA technique for the bert-base-multilingual-cased model in Approach 3 (a) and Approach (4) of a representative set of sentence pairs from the STS test split with different levels of semantic similarity.}
\label{fig:bert-base-visu-example}
\end{figure}

\subsection{Dimensionality reduction in fine-tuned models}

Although it was stated at the beginning of this section that model comparison is not an objective of the paper, different versions of the same architecture have been included for a wider evaluation of the effects of dimensionality reduction techniques (i.e., \textit{xlm-roberta-base} \& \textit{xlm-roberta-large}, \textit{bert-base-multilingual-cased} \& \textit{distilbert-base-multilingual-cased}) have been included in the study. Based on the complexity and learning potential of the models, one would expect the \textit{xlm-roberta-large} model to perform better than the \textit{xlm-roberta-base }model. Similarly, the \textit{bert-base-multilingual-cased} model would be expected to be superior to the \textit{distilbert-base-multilingual-cased} model. In Approach 1, however, the opposite is true. Only when fine-tuning occurs in the task (Approaches 2 and 4) is it observed how the performance of the models is in line with the expected complexity (see Table \ref{table:best-tech-ap4}). These results provide wider support for the importance of supervised fine-tuning.


Similarly, fine-tuning also alters the impact of dimensionality reduction techniques on the results of multilingual models. Compared to Approach 3, when fine-tuning, the feature selection techniques and nonlinearity becomes more important, ICA becomes less critical, and the minimum number of dimensions that outperform the baseline approach increases. 

As can be seen in Figure \ref{fig:dim-red-plot-ap4} and Table \ref{table:best-tech-ap4}, the promotion of Variance Threshold feature selection as one of the best techniques for some models in Approach 4 could be attributed to the fact that fine-tuning adjust the embeddings to the task, reducing the presence of noisy variable and taking advantage of the variable selection process. This would be in line with the results obtained in Approach 3, where feature extraction techniques more adequately handled the presence of unadjusted variables. This further supports the argument that they can reduce dimensions and generate new feature representations to help improve performance on learning problems.

Furthermore, Table \ref{table:dim-red-fast-thres-ap4} shows the lack of ICA dominance and the emergence of the KPCA-sigmoid technique as the method that with fewer dimensions improves the baseline Approach 2 ($\SI{59.32 \pm 29.92}{\percent}$ reduced dimensions retaining $\SI{99.00 \pm 2.00}{\percent}$ baseline's performance). It reveals that managing the non-Gaussianity issue is less relevant than the nonlinearity issue after fine-tuning. The fine-tuning process also impacts the reduction capabilities of the dimensionality reduction techniques since considering the technique that retains the maximum performance with the lowest dimensions for each model in Table \ref{table:dim-red-fast-thres-ap4} show that the initial dimensions from the baseline Approach 2 are reduced an average of $\SI{54.65 \pm 32.20}{\percent}$. Although this average reduction is lower than the achieved earlier in the comparison of Approach 3 with the baseline Approach 1, it still remarkably that even after fine-tuning, the multilingual performance can be exceeded with half of the dimensions. Finally, the two-tailed paired T-test comparing Approach 2 vs Approach 4 using the values from Table \ref{table:best-tech-ap4} resulted in $p = 0.255$, revealing that there is no significant difference in performance using these dimensional reduction techniques.

Finally, we shall conclude by analyzing the results obtained with UMAP. The case of UMAP shows that for the STS task, it is not a suitable technique to reduce the dimensionality of the embeddings since it is the one that retains the lowest percentage of the baseline performances in both pre-trained and fine-tuned embeddings. Considering this fact, it is interesting to note that the potential of this technique resides in the fact that it quickly saturates, i.e., the maximum retained performance is reached with a small number of dimensions in Approach 3 with an average of $\SI{94.65 \pm 6.07}{\percent}$ of reduced initial dimensions retaining $\SI{49.00 \pm 11.14}{\percent}$ performance with respect to the reference Approach 1, and more notably in Approach 4 with an average of $\SI{98.42 \pm 0.72}{\percent}$ of reduced initial dimensions retaining $\SI{76.00 \pm 3.74}{\percent}$ performance with respect to the reference Approach 2.

\subsection{Visualization capabilities}

\newcolumntype{L}{>{\RaggedRight\arraybackslash}X}
\newcolumntype{Y}{>{\centering\arraybackslash}X}

\begin{table}[htpb]
\centering
\caption{Analysis of the information retained for 2D and 3D embeddings visualization from Approach 3.}
\label{table:visu-ap3}
\def\arraystretch{1.25}%
\begin{tabular}{clccc} 
\toprule
\textbf{Model}                                                                                         & \textbf{Technique}    & \textbf{Avg $r_s$ 2D} & \textbf{Avg $r_s$ 3D}  \\ 
\hline
\multirow{8}{*}{\begin{tabular}[c]{@{}c@{}}bert-base-\\multilingual-cased\end{tabular}}       & IPCA         &     0.0269      & 0.0518         \\
                                                                                              & ICA          &      0.0657      &   0.1064        \\
                                                                                              &  poly    &     0.0516      &    0.0850      \\
                                                                                              &  rbf     &     0.0577       &    0.0825       \\
                                                                                              &  sigmoid &     0.0524	       & 0.0849       \\
                                                                                              &  cosine  &      0.0479     &    0.0792      \\
                                                                                              & UMAP         &     \textbf{ 0.0738}      &    \textbf{0.1578}      \\
                                                                                              & VarThres     &     -       &    -      \\ 
\hline
\multirow{8}{*}{\begin{tabular}[c]{@{}c@{}}distilbert-base-\\multilingual-cased\end{tabular}} & IPCA         &     0.0313       &  0.0697          \\
                                                                                              & ICA          &     0.0431       &    0.0766       \\
                                                                                              &  poly    &      0.0359     &       0.0705    \\
                                                                                              &  rbf     &       \textbf{0.1120 }    &    0.0776       \\
                                                                                              &  sigmoid &      0.0567      &      0.0769     \\
                                                                                              &  cosine  &       0.0466     &    0.0762      \\
                                                                                              & UMAP         &      0.0865      &   \textbf{ 0.1431}       \\
                                                                                              & VarThres     &     -       &  -         \\ 
\hline
\multirow{8}{*}{xlm-roberta-base     }                                                      & IPCA         &  0.0756          &  0.0933         \\
                                                                                              & ICA          &    \textbf{ 0.1117 }     &    \textbf{0.1665 }      \\
                                                                                              &  poly    &      0.0922      &      0.1047     \\
                                                                                              &  rbf     &       0.0811     &    0.1086       \\
                                                                                              &  sigmoid &      0.0767       &   0.0986       \\
                                                                                              &  cosine  &      0.0758      &      0.0963    \\
                                                                                              & UMAP         &      0.0652    &    0.0855       \\
                                                                                              & VarThres     &       -     &      -    \\ 
\hline
\multirow{8}{*}{xlm-roberta-large}                                                            & IPCA         &   0.0688       &     \textbf{0.0992}      \\
                                                                                              & ICA          &     \textbf{0.0845}       &   \textbf{ 0.0992}       \\
                                                                                              &  poly    &      0.0612     &      0.0857     \\
                                                                                              &  rbf     &       0.0616
     &    0.0939       \\
                                                                                              &  sigmoid &     0.0656      &    0.0827       \\
                                                                                              &  cosine  &     0.0825      &      0.0910     \\
                                                                                              & UMAP         &       0.0613     &    0.0950       \\
                                                                                              & VarThres     &      -      &       -    \\
\hline
\multirow{8}{*}{LaBSE}                                                                        & IPCA         &    0.1972       &     0.3107      \\
                                                                                              & ICA          &     0.1659       &    0.2509       \\
                                                                                              &  poly    &     \textbf{0.2016}     &     \textbf{ 0.3175}   \\
                                                                                              &  rbf     &       0.1991
     &    0.2513      \\
                                                                                              &  sigmoid &     0.1885      &    0.2380      \\
                                                                                              &  cosine  &    0.1783      &      0.2230     \\
                                                                                              & UMAP         &       0.1710    &   0.2944      \\
                                                                                              & VarThres     &      -      &       -    \\
\bottomrule
\end{tabular}
\end{table}
\newcolumntype{L}{>{\RaggedRight\arraybackslash}X}
\newcolumntype{Y}{>{\centering\arraybackslash}X}

\begin{table}[htpb]
\centering
\caption{Analysis of the information retained for 2D and 3D embeddings visualization from Approach 4.}
\def\arraystretch{1.25}%
\begin{tabular}{clccc} 
\toprule
\begin{tabular}[c]{@{}l@{}}\textbf{Model}\end{tabular}                                                   & \textbf{Technique} & \textbf{Avg $r_s$ 2D} & \textbf{Avg $r_s$ 3D}  \\ 
\hline
\multirow{8}{*}{\begin{tabular}[c]{@{}c@{}}bert-base-\\multilingual-cased-\\fine-tuned\end{tabular}}       & IPCA               & \textbf{0.3481}       & 0.4190                 \\
                                                                                                           & ICA                & 0.3170                & 0.4244                 \\
                                                                                                           &  poly          & 0.3431                & 0.3948                 \\
                                                                                                           &  rbf           & 0.3027                & 0.4012                 \\
                                                                                                           &  sigmoid       & 0.3176                & \textbf{ 0.4237 }      \\
                                                                                                           &  cosine        & 0.3120                & 0.4218                 \\
                                                                                                           & UMAP               & 0.2961                & 0.3858                 \\
                                                                                                           & VarThres           & -                     & -                      \\ 
\hline
\multirow{8}{*}{\begin{tabular}[c]{@{}c@{}}distilbert-base-\\multilingual-cased-\\fine-tuned\end{tabular}} & IPCA               & 0.2908                & 0.3692                 \\
                                                                                                           & ICA                & 0.2650                & 0.3755                 \\
                                                                                                           &  poly          & 0.2636                & 0.3660                 \\
                                                                                                           &  rbf           & 0.2689                & 0.3741                 \\
                                                                                                           &  sigmoid       & 0.2667                & 0.3777                 \\
                                                                                                           &  cosine        & 0.2624                & 0.3733                 \\
                                                                                                           & UMAP               & \textbf{0.3322}       & \textbf{ 0.4256}       \\
                                                                                                           & VarThres           & -                     & -                      \\ 
\hline
\multirow{8}{*}{\begin{tabular}[c]{@{}c@{}}xlm-roberta-base\\-fine-tuned\end{tabular}}                     & IPCA               & \textbf{0.3690}       & \textbf{0.4406}        \\
                                                                                                           & ICA                & 0.3683                & 0.4357                 \\
                                                                                                           &  poly          & 0.3644                & 0.4347                 \\
                                                                                                           &  rbf           & 0.3627                & 0.4334                 \\
                                                                                                           &  sigmoid       & 0.3619                & 0.4319                 \\
                                                                                                           &  cosine        & 0.3635                & 0.4333                 \\
                                                                                                           & UMAP               & 0.2809                & 0.3560                 \\
                                                                                                           & VarThres           & -                     & -                      \\ 
\hline
\multirow{8}{*}{\begin{tabular}[c]{@{}c@{}}xlm-roberta-large\\-fine-tuned\end{tabular}}                    & IPCA               & 0.3763                & \textbf{ 0.4574 }      \\
                                                                                                           & ICA                & 0.3306                & 0.4411                 \\
                                                                                                           &  poly          & 0.3779                & 0.4517                 \\
                                                                                                           &  rbf           & 0.3540                & 0.4435                 \\
                                                                                                           &  sigmoid       & 0.3798                & 0.4510                 \\
                                                                                                           &  cosine        & 0.3792                & 0.4503                 \\
                                                                                                           & UMAP               & \textbf{0.3959 }      & 0.4510                 \\
                                                                                                           & VarThres           & 0.3526                & -                      \\ 
\hline
\multirow{8}{*}{LaBSE-fine-tuned}                                                                          & IPCA               &\textbf{ 0.3311}                & 0.4153                 \\
                                                                                                           & ICA                & 0.3277      & 0.4130                 \\
                                                                                                           &  poly          & 0.3274                & \textbf{0.4205}                 \\
                                                                                                           &  rbf           & 0.2842                & 0.3452                 \\
                                                                                                           &  sigmoid       & 0.3272                & 0.4167        \\
                                                                                                           &  cosine        & 0.3160                & 0.4018                 \\
                                                                                                           & UMAP               & 0.2192                & 0.4103                 \\
                                                                                                           & VarThres           & -                     & -                      \\
\bottomrule
\end{tabular}

\label{table:visu-ap4}
\end{table}

In addition to embedding compression and better generalization capabilities, one of the primary uses of dimensionality reduction techniques is facilitating data visualization. In NLP, data visualization can be helpful to interpret the model learning process, semantic capabilities and its results. For instance, Mikolov et al.~\cite{mikolov2013efficient} explored the projections of high dimensional word embeddings in 2D to extract multiple relationships between words. For this reason, in this work we have also applied the different techniques in 2D and 3D dimension reduction to study their usefulness for the visualization and model's interpretability.

A straightforward example of the applicability of these techniques can be found in Figure~\ref{fig:bert-base-visu-example}. In this Figure, we compare the 2D embeddings representation of a random subset pairs of sentences of the STS dataset in the English language with different levels of semantic similarity. The comparison between Figure~\ref{fig:bert-base-visu-example-pretrain} and Figure~\ref{fig:bert-base-visu-example-fine-tuned} allows us to observe the effect of fine-tuning in the \textit{bert-base-multilingual-cased model}. In the pre-trained case, we observe how the distance between pairs of sentences is similar and does not match the labelled similarity. However, after fine-tuning, we observed a better agreement between the spatial representation and the labelled similarity, demonstrating and supporting the improvement of downstream task fine-tuning.

Besides, the numerical results of these analysis are presented in Table \ref{table:visu-ap3} for Approach 3 and Table \ref{table:visu-ap4} for Approach 4. In both cases there is no clear dominance of any technique. The most likely explanation of this variety of results is that dimensionality reduction for visualization purposes is highly model-dependent. As anticipated, feature extraction techniques clearly prove to be more useful than feature selection in both approaches, since transforming the initial high-dimensional space into a new reduced latent space can extract more information rather than selecting specific variables, even if these variables are properly adjusted to the task.

Regarding UMAP and visualization, it does not show noteworthy results either. However, it is intriguing to note that it turns out to be the best technique for visualization in some models, which corroborates the saturation capacity hypothesis observed in the previous analyses. The discrepancy of the best visualization technique between models could be attributed to the previous mentioned model dependency. Another possible explanation for this might be that the greatest potential of this technique resides precisely in visualization, as shown by other works  where this capability is exploited, such as in Health Science for identifying cell types~\cite{umap_app_2019_nature} or in Artificial Intelligence to analyze the activations throughout modern convolutional neural networks~\cite{umap_carter2019activation}. However, we assume the limitations that may derive from the task and the models used in this study, given that the technique is highly dependent on the parameters used, which unfortunately were not possible to be intensively explored.

Altogether, it is advisable to explore all techniques for visualisation purposes and select the one that best suits the desired task.

\section{Conclusion}

In this investigation, the goal was to assess the impact of a variety of dimensionality reduction techniques on the performance of pre-computed multilingual siamese fashion Transformers embeddings on semantic textual similarity tasks from mSTSb, to expand our knowledge of semantic-aware transformer-based models. To this end, two different baseline approaches are reduced (i.e., Approach 1 and Approach 2), one using the pre-trained version of the models and the second one further fine-tuning them on the downstream STS task. Particular attention is paid to analysing which techniques best and with the fewest dimensions improve the performance of the baseline approaches.


From the research carried out, it is possible to conclude that dimensionality reduction techniques can help to reduce the number of dimensions of the embeddings while improving the results if using pre-trained embeddings from Approach 1 and preserving the performance when using fine-tuned embeddings from Approach 2. Nevertheless, the dimensionality reduction is more considerable in the pre-trained version with an average of  $\SI{91.58 \pm 2.59}{\percent}$ compared to the average $\SI{54.65 \pm 32.20}{\percent}$ of fine-tuned version. Special attention is given to ICA in the pre-trained scenario, which proved to manage adequately the noisy variables present in not adjusted embeddings. This technique also proved to be a reasonable alternative to fit the models in the downstream task in an unsupervised way, leading to a generalized adjusted version of the models with downstream multitasking capabilities. Nevertheless, it has also been proved that this unsupervised fitting is not comparable to a supervised fine-tuning. On the other hand, the fine-tuned scenario revealed the relevance of feature selection techniques and the significance of nonlinear KPCA techniques for dimensionality reduction.

Our experiments results are consistent with previous results obtained by Raunak et al.~\cite{explore-embs-dim-red-2019, RaunakVikas2019ODLP} corroborating the hypothesis that embeddings reduction can maintain or improve the performance of the original embeddings by extending their evaluation in state-of-the-art contextual-based models from a multilingual approach. In this way, we can establish that dimensionality reduction techniques could be leveraged for contextualized embeddings as well.

This work has also considered the consequences of dimensionality reduction for visualization purposes by extending previous work in the literature~\cite{visu-dim-red-2021}. The results corroborate that feature extraction methods are more practical for STS tasks than feature selection. However, selecting the best dimensionality reduction technique for visualization depends on the model, so it is recommended to explore different techniques for this purpose.

To our knowledge, this is the first study to investigate the effect of dimensionality reduction techniques and Transformers models in a multilingual semantic-awareness scenario. Based on the promising findings presented in this paper, continued research into the impact of dimensionality reduction techniques in other high-demanded NLP tasks appears totally justified. Furthermore, in future research, we intend to concentrate on testing the reduced models presented in this work in real-world applications. As previously stated, these findings of multilingual semantic similarity are of direct practical applicability. The combination of dimensionality reduction techniques with Transformers models could also help reduce the embeddings size and to make possible ensemble approaches. Finally, further studies about the multitasking generalization capabilities of ICA for pre-trained models are still required.

\label{sec:conclusion}

\section*{Acknowledgements}
This research has been supported by the Spanish Ministry of Science and Education under FightDIS (PID2020-117263GB-100) and XAI-Disinfodemics (PLEC2021-007681) grants, by Comunidad Aut\'{o}noma de Madrid under S2018/ TCS-4566 (CYNAMON) grant, by BBVA Foundation grants for scientific research teams SARS-CoV-2 and COVID-19 under the grant: "\textit{CIVIC: Intelligent characterisation of the veracity of the information related to COVID-19}", and by IBERIFIER (Iberian Digital Media Research and Fact-Checking Hub), funded by the European Commission under the call CEF-TC-2020-2, grant number 2020-EU-IA-0252. Finally, David Camacho has been supported by the Comunidad Aut\'{o}noma de Madrid under: "Convenio Plurianual with the Universidad Politécnica de Madrid in the actuation line of \textit{Programa de Excelencia para el Profesorado Universitario}".

\bibliographystyle{IEEEtran}
\bibliography{references}

\begin{thebibliography}{10}
\providecommand{\url}[1]{#1}
\csname url@samestyle\endcsname
\providecommand{\newblock}{\relax}
\providecommand{\bibinfo}[2]{#2}
\providecommand{\BIBentrySTDinterwordspacing}{\spaceskip=0pt\relax}
\providecommand{\BIBentryALTinterwordstretchfactor}{4}
\providecommand{\BIBentryALTinterwordspacing}{\spaceskip=\fontdimen2\font plus
\BIBentryALTinterwordstretchfactor\fontdimen3\font minus
  \fontdimen4\font\relax}
\providecommand{\BIBforeignlanguage}[2]{{%
\expandafter\ifx\csname l@#1\endcsname\relax
\typeout{** WARNING: IEEEtran.bst: No hyphenation pattern has been}%
\typeout{** loaded for the language `#1'. Using the pattern for}%
\typeout{** the default language instead.}%
\else
\language=\csname l@#1\endcsname
\fi
#2}}
\providecommand{\BIBdecl}{\relax}
\BIBdecl

\bibitem{OtterDanielW2021ASot}
D.~W. Otter, J.~R. Medina, and J.~K. Kalita, ``\BIBforeignlanguage{eng}{A
  survey of the usages of deep learning for natural language processing},''
  \emph{\BIBforeignlanguage{eng}{IEEE transaction on neural networks and
  learning systems}}, vol.~32, no.~2, pp. 604--624, 2021.

\bibitem{TayYi2020ETAS}
Y.~Tay, M.~Dehghani, D.~Bahri, and D.~Metzler, ``Efficient transformers: A
  survey,'' 2020.

\bibitem{vaswani2017attention}
A.~Vaswani, N.~Shazeer, N.~Parmar, J.~Uszkoreit, L.~Jones, A.~N. Gomez,
  L.~Kaiser, and I.~Polosukhin, ``Attention is all you need,'' 2017.

\bibitem{devlin2018bert}
J.~Devlin, M.-W. Chang, K.~Lee, and K.~Toutanova, ``Bert: Pre-training of deep
  bidirectional transformers for language understanding,'' \emph{arXiv preprint
  arXiv:1810.04805}, 2018.

\bibitem{reimers2019sentencebert}
N.~Reimers and I.~Gurevych, ``Sentence-bert: Sentence embeddings using siamese
  bert-networks,'' 2019.

\bibitem{bertuit_2022}
J.~Huertas-Tato, A.~Martin, and D.~Camacho, ``Bertuit: Understanding spanish
  language in twitter through a native transformer,'' 2022.

\bibitem{NLP-chowdhury}
G.~G. Chowdhury, ``Natural language processing,'' \emph{Annual Review of
  Information Science and Technology}, vol.~37, no.~1, pp. 51--89, 2003.

\bibitem{Cer_2017}
D.~Cer, M.~Diab, E.~Agirre, I.~Lopez-Gazpio, and L.~Specia, ``Semeval-2017 task
  1: Semantic textual similarity multilingual and crosslingual focused
  evaluation,'' \emph{Proceedings of the 11th International Workshop on
  Semantic Evaluation (SemEval-2017)}, 2017.

\bibitem{humeau2020polyencoders}
S.~Humeau, K.~Shuster, M.-A. Lachaux, and J.~Weston, ``Poly-encoders:
  Transformer architectures and pre-training strategies for fast and accurate
  multi-sentence scoring,'' 2020.

\bibitem{zhelezniak2019correlation}
V.~Zhelezniak, A.~Savkov, A.~Shen, and N.~Y. Hammerla, ``Correlation
  coefficients and semantic textual similarity,'' 2019.

\bibitem{SIDOROV2014}
G.~Sidorov, A.~Gelbukh, H.~Gómez-Adorno, and D.~Pinto, ``Soft similarity and
  soft cosine measure: similarity of features in vector space model,''
  \emph{Computación y Sistemas}, vol.~18, no.~3, pp. 491 -- 504, 2014.

\bibitem{AraqueOscar2017Edls}
O.~Araque, I.~Corcuera-Platas, J.~F. Sánchez-Rada, and C.~A. Iglesias,
  ``\BIBforeignlanguage{eng}{Enhancing deep learning sentiment analysis with
  ensemble techniques in social applications},''
  \emph{\BIBforeignlanguage{eng}{Expert systems with applications}}, vol.~77,
  pp. 236--246, 2017.

\bibitem{ensemble-fake}
Y.~Zhou, Y.~Yang, H.~Liu, X.~Liu, and N.~Savage, ``Deep learning based fusion
  approach for hate speech detection,'' \emph{IEEE Access}, vol.~8, pp.
  128\,923--128\,929, 2020.

\bibitem{KhanAsifullah2020Asot}
A.~Khan, A.~Sohail, U.~Zahoora, and A.~S. Qureshi, ``\BIBforeignlanguage{eng}{A
  survey of the recent architectures of deep convolutional neural networks},''
  \emph{\BIBforeignlanguage{eng}{The Artificial intelligence review}}, vol.~53,
  no.~8, pp. 5455--5516, 2020.

\bibitem{semeval-2020-semantic}
\BIBentryALTinterwordspacing
A.~Herbelot, X.~Zhu, A.~Palmer, N.~Schneider, J.~May, and E.~Shutova, Eds.,
  \emph{Proceedings of the Fourteenth Workshop on Semantic Evaluation}.\hskip
  1em plus 0.5em minus 0.4em\relax Barcelona (online): International Committee
  for Computational Linguistics, Dec. 2020. [Online]. Available:
  \url{https://www.aclweb.org/anthology/2020.semeval-1.0}
\BIBentrySTDinterwordspacing

\bibitem{10.1007/978-3-030-28577-7_1}
N.~Ferro, ``What happened in {CLEF}{\ldots} {F}or a while?'' in
  \emph{Experimental IR Meets Multilinguality, Multimodality, and Interaction},
  F.~Crestani, M.~Braschler, J.~Savoy, A.~Rauber, H.~M{\"u}ller, D.~E. Losada,
  G.~Heinatz~B{\"u}rki, L.~Cappellato, and N.~Ferro, Eds.\hskip 1em plus 0.5em
  minus 0.4em\relax Cham: Springer International Publishing, 2019, pp. 3--45.

\bibitem{hyperspectral_2019}
W.~Sun and Q.~Du, ``{Hyperspectral Band Selection A review},'' \emph{{IEEE
  GEOSCIENCE AND REMOTE SENSING MAGAZINE}}, vol.~{7}, no.~{2}, pp. {118--139},
  {JUN} {2019}.

\bibitem{review_ft_select_2020}
S.~Solorio-Fernandez, J.~A. Carrasco-Ochoa, and J.~F. Martinez-Trinidad, ``{A
  review of unsupervised feature selection methods},'' \emph{{ARTIFICIAL
  INTELLIGENCE REVIEW}}, vol.~{53}, no.~{2}, pp. {907--948}, {FEB} {2020}.

\bibitem{app_ft_ext_slect_2018}
A.~E. Maxwell, T.~A. Warner, and F.~Fang, ``{Implementation of machine-learning
  classification in remote sensing: an applied review},'' \emph{{INTERNATIONAL
  JOURNAL OF REMOTE SENSING}}, vol.~{39}, no.~{9}, pp. {2784--2817}, {2018}.

\bibitem{PatelAnkurA2019Hulu}
A.~A. Patel, \emph{\BIBforeignlanguage{eng}{Hands-on unsupervised learning
  using Python : how to build applied machine learning solutions from unlabeled
  data}}.\hskip 1em plus 0.5em minus 0.4em\relax Sebastopol, California:
  O'Reilly, 2019.

\bibitem{HiraZenaM2015ARoF}
Z.~M. Hira and D.~F. Gillies, ``\BIBforeignlanguage{eng}{A review of feature
  selection and feature extraction methods applied on microarray data},''
  \emph{\BIBforeignlanguage{eng}{Advances in bioinformatics}}, vol. 2015, pp.
  198\,363--13, 2015.

\bibitem{pretrain_fine_tune_paradigma_2021}
D.~Xu, I.~E. Yen, J.~Zhao, and Z.~Xiao, ``Rethinking network pruning - under
  the pre-train and fine-tune paradigm,'' \emph{CoRR}, vol. abs/2104.08682,
  2021.

\bibitem{on_the_fly_embeddings_2017}
D.~Bahdanau, T.~Bosc, S.~Jastrzebski, E.~Grefenstette, P.~Vincent, and
  Y.~Bengio, ``Learning to compute word embeddings on the fly,'' \emph{CoRR},
  vol. abs/1706.00286, 2017.

\bibitem{johnson2019billion}
J.~Johnson, M.~Douze, and H.~J{\'e}gou, ``Billion-scale similarity search with
  {GPUs},'' \emph{IEEE Transactions on Big Data}, vol.~7, no.~3, pp. 535--547,
  2019.

\bibitem{explore-embs-dim-red-2019}
V.~Raunak, V.~Gupta, and F.~Metze, ``Effective dimensionality reduction for
  word embeddings,'' in \emph{4TH WORKSHOP ON REPRESENTATION LEARNING FOR NLP
  (REPL4NLP-2019)}.\hskip 1em plus 0.5em minus 0.4em\relax {Facebook AI Res;
  Amazon; Naver Labs Europe; Assoc Computat Linguist, Special Interest Grp
  Representat Learning}, {2019}, pp. {235--243}, {4th Workshop on
  Representation Learning for NLP (RepL4NLP), Florence, ITALY, AUG 02, 2019}.

\bibitem{RaunakVikas2019ODLP}
V.~Raunak, V.~Kumar, V.~Gupta, and F.~Metze, ``On dimensional linguistic
  properties of the word embedding space,'' \emph{CoRR}, vol. abs/1910.02211,
  2019.

\bibitem{maria_mihaela_post-processing_2021}
T.~Maria~Mihaela, A.~Anamaria, G.~Simona~Elena, and A.~Crisan,
  ``Post-{Processing} and {Dimensionality} {Reduction} for {Extreme} {Learning}
  {Machine} in {Text} {Classification},'' \emph{ECONOMIC COMPUTATION AND
  ECONOMIC CYBERNETICS STUDIES AND RESEARCH}, vol.~55, no. 4/2021, pp. 37--50,
  Dec. 2021.

\bibitem{mitra_introduction_2018}
B.~Mitra and N.~Craswell, ``\BIBforeignlanguage{English}{An {Introduction} to
  {Neural} {Information} {Retrieval}},''
  \emph{\BIBforeignlanguage{English}{Foundations and Trends® in Information
  Retrieval}}, vol.~13, no.~1, pp. 1--126, Dec. 2018.

\bibitem{Camastra2008}
F.~Camastra and A.~Vinciarelli, ``\BIBforeignlanguage{en}{Feature {Extraction}
  {Methods} and {Manifold} {Learning} {Methods}},'' in
  \emph{\BIBforeignlanguage{en}{Machine {Learning} for {Audio}, {Image} and
  {Video} {Analysis}}}.\hskip 1em plus 0.5em minus 0.4em\relax London: Springer
  London, 2008, pp. 305--341.

\bibitem{CO2_machine_learning_2019}
E.~Strubell, A.~Ganesh, and A.~McCallum, ``{Energy and Policy Considerations
  for Deep Learning in NLP},'' in \emph{{57TH ANNUAL MEETING OF THE ASSOCIATION
  FOR COMPUTATIONAL LINGUISTICS (ACL 2019)}}, {Korhonen, A and Traum, D and
  Marquez, L}, Ed.\hskip 1em plus 0.5em minus 0.4em\relax {Assoc Computat
  Linguist; Apple; ASAPP; Bloomberg Engn; Bosch; Expedia; Facebook; Google;
  Microsoft; Salesforce; Amazon; Baidu; DeepMind; Grammarly; Huawei; IBM;
  Tencent; ByteDance; DiDi; Keiosk Analyt; Megagon Labs; Naver; PolyAi;
  Samsung; Bebelscape; BMW; Cisco; Duolingo; Ebay; G Res; SAP; Raytheon BBN
  Technologies; USC Viterbi, Sch Engn; Shannon Ai}, {2019}, pp. {3645--3650},
  {57th Annual Meeting of the Association-for-Computational-Linguistics (ACL),
  Florence, ITALY, JUL 28-AUG 02, 2019}.

\bibitem{JDH17}
J.~Johnson, M.~Douze, and H.~J{\'e}gou, ``Billion-scale similarity search with
  gpus,'' \emph{arXiv preprint arXiv:1702.08734}, 2017.

\bibitem{pennington2014GloVe}
\BIBentryALTinterwordspacing
J.~Pennington, R.~Socher, and C.~D. Manning, ``Glove: Global vectors for word
  representation,'' in \emph{Empirical Methods in Natural Language Processing
  (EMNLP)}, 2014, pp. 1532--1543. [Online]. Available:
  \url{http://www.aclweb.org/anthology/D14-1162}
\BIBentrySTDinterwordspacing

\bibitem{bojanowski2017fastext}
P.~Bojanowski, E.~Grave, A.~Joulin, and T.~Mikolov, ``Enriching word vectors
  with subword information,'' \emph{Transactions of the Association for
  Computational Linguistics}, vol.~5, pp. 135--146, 2017.

\bibitem{pca-original-1901}
K.~Pearson, ``\BIBforeignlanguage{en}{{LIII}. \textit{{On} lines and planes of
  closest fit to systems of points in space}},''
  \emph{\BIBforeignlanguage{en}{The London, Edinburgh, and Dublin Philosophical
  Magazine and Journal of Science}}, vol.~2, no.~11, pp. 559--572, Nov. 1901.

\bibitem{PCA_review}
I.~T. Jolliffe and J.~Cadima, ``{Principal component analysis: a review and
  recent developments},'' \emph{{PHILOSOPHICAL TRANSACTIONS OF THE ROYAL
  SOCIETY A-MATHEMATICAL PHYSICAL AND ENGINEERING SCIENCES}}, vol. {374}, no.
  {2065}, {APR 13} {2016}.

\bibitem{LSA_DeerwesterScott1990}
S.~Deerwester, S.~T. Dumais, G.~W. Furnas, T.~K. Landauer, and R.~Harshman,
  ``\BIBforeignlanguage{eng}{Indexing by latent semantic analysis},''
  \emph{\BIBforeignlanguage{eng}{Journal of the American Society for
  Information Science}}, vol.~41, no.~6, pp. 391--407, 1990.

\bibitem{ShimomotoEricaK2021Tcbo}
E.~K. Shimomoto, F.~Portet, and K.~Fukui, ``\BIBforeignlanguage{eng}{Text
  classification based on the word subspace representation},''
  \emph{\BIBforeignlanguage{eng}{Pattern analysis and applications : PAA}},
  vol.~24, no.~3, pp. 1075--1093, 2021.

\bibitem{SongHaohao2020ECwR}
H.~Song, D.~Zou, L.~Hu, and J.~Yuan, ``\BIBforeignlanguage{eng}{Embedding
  compression with right triangle similarity transformations},'' in
  \emph{\BIBforeignlanguage{eng}{Artificial Neural Networks and Machine
  Learning – ICANN 2020}}, ser. Lecture Notes in Computer Science.\hskip 1em
  plus 0.5em minus 0.4em\relax Cham: Springer International Publishing, 2020,
  pp. 773--785.

\bibitem{visu-dim-red-2021}
R.~Choudhary, S.~Doboli, and A.~A. Minai, ``A comparative study of methods for
  visualizable semantic embedding of small text corpora,'' in \emph{2021
  International Joint Conference on Neural Networks (IJCNN)}, 2021, pp. 1--8.

\bibitem{Hinton2002StochasticNE}
G.~Hinton and S.~Roweis, ``Stochastic neighbor embedding,'' in
  \emph{Proceedings of the 15th International Conference on Neural Information
  Processing Systems}, ser. NIPS'02.\hskip 1em plus 0.5em minus 0.4em\relax
  Cambridge, MA, USA: MIT Press, 2002, p. 857–864.

\bibitem{de_rosa_how_2019}
G.~H. de~Rosa, J.~R.~F. Brega, and J.~P. Papa, ``\BIBforeignlanguage{en}{How
  optimizing perplexity can affect the dimensionality reduction on word
  embeddings visualization?}'' \emph{\BIBforeignlanguage{en}{SN Applied
  Sciences}}, vol.~1, no.~12, p. 1673, Dec. 2019.

\bibitem{dim_red_text_2013}
Y.~Bengio, A.~Courville, and P.~Vincent, ``Representation learning: A review
  and new perspectives,'' \emph{IEEE Transactions on Pattern Analysis and
  Machine Intelligence}, vol.~35, no.~8, pp. 1798--1828, 2013.

\bibitem{LiuYimin20213CAd}
Y.~Liu and L.~J. Durlofsky, ``\BIBforeignlanguage{eng}{3d cnn-pca: A
  deep-learning-based parameterization for complex geomodels},''
  \emph{\BIBforeignlanguage{eng}{Computers \& geosciences}}, vol. 148, p.
  104676, 2021.

\bibitem{ChoiSangWon2021APtd}
S.~W. Choi and B.~H.~S. Kim, ``\BIBforeignlanguage{eng}{Applying pca to deep
  learning forecasting models for predicting pm2.5},''
  \emph{\BIBforeignlanguage{eng}{Sustainability (Basel, Switzerland)}},
  vol.~13, no.~7, p. 3726, 2021.

\bibitem{text_dim_red_app_2020}
N.~Kushwaha and M.~Pant, ``Textual data dimensionality reduction - a deep
  learning approach,'' \emph{Multimedia Tools Appl.}, vol.~79, no. 15–16, p.
  11039–11050, apr 2020.

\bibitem{bio_dnn_dim_app_2021}
D.~Menaga and S.~Revathi, ``Probabilistic principal component analysis (ppca)
  based dimensionality reduction and deep learning for cancer classification,''
  in \emph{Intelligent Computing and Applications}, S.~S. Dash, S.~Das, and
  B.~K. Panigrahi, Eds.\hskip 1em plus 0.5em minus 0.4em\relax Singapore:
  Springer Singapore, 2021, pp. 353--368.

\bibitem{nogueira2020document}
R.~Nogueira, Z.~Jiang, and J.~Lin, ``Document ranking with a pretrained
  sequence-to-sequence model,'' 2020.

\bibitem{mSTSb-ref}
{\'A}.~Huertas-Garc{\'i}a, J.~Huertas-Tato, A.~Mart{\'i}n, and D.~Camacho,
  ``Countering misinformation through semantic-aware multilingual models,'' in
  \emph{Intelligent Data Engineering and Automated Learning -- IDEAL 2021},
  H.~Yin, D.~Camacho, P.~Tino, R.~Allmendinger, A.~J. Tall{\'o}n-Ballesteros,
  K.~Tang, S.-B. Cho, P.~Novais, and S.~Nascimento, Eds.\hskip 1em plus 0.5em
  minus 0.4em\relax Cham: Springer International Publishing, 2021, pp.
  312--323.

\bibitem{bm25}
S.~Robertson, H.~Zaragoza, and M.~Taylor, ``Simple bm25 extension to multiple
  weighted fields,'' in \emph{Proceedings of the Thirteenth ACM International
  Conference on Information and Knowledge Management}, ser. CIKM '04.\hskip 1em
  plus 0.5em minus 0.4em\relax New York, NY, USA: Association for Computing
  Machinery, 2004, p. 42–49.

\bibitem{wardle_information_2017}
\BIBentryALTinterwordspacing
C.~Wardle and H.~Derakhshan, ``Information disorder: {Toward} an
  interdisciplinary framework for research and policy making,'' Council of
  Europe, Tech. Rep., Oct. 2017. [Online]. Available:
  \url{https://rm.coe.int/information-disorder-toward-an-interdisciplinary-frameworkfor-researc/168076277c}
\BIBentrySTDinterwordspacing

\bibitem{CarmiElinor2020DcRd}
E.~Carmi, S.~J. Yates, E.~Lockley, and A.~Pawluczuk,
  ``\BIBforeignlanguage{eng}{Data citizenship: Rethinking data literacy in the
  age of disinformation, misinformation, and malinformation},''
  \emph{\BIBforeignlanguage{eng}{Internet policy review}}, vol.~9, no.~2, 2020.

\bibitem{9120902}
J.~{Gaglani}, Y.~{Gandhi}, S.~{Gogate}, and A.~{Halbe}, ``Unsupervised whatsapp
  fake news detection using semantic search,'' in \emph{2020 4th International
  Conference on Intelligent Computing and Control Systems (ICICCS)}, 2020, pp.
  285--289.

\bibitem{huertas2021civic}
\BIBentryALTinterwordspacing
{\'{A}}.~Huertas{-}Garc{\'{\i}}a, J.~Huertas{-}Tato, A.~Mart{\'{\i}}n, and
  D.~Camacho, ``{CIVIC-UPM} at checkthat!~2021: Integration of transformers in
  misinformation detection and topic classification,'' in \emph{Proceedings of
  the Working Notes of {CLEF} 2021 - Conference and Labs of the Evaluation
  Forum, Bucharest, Romania, September 21st - to - 24th, 2021}, ser. {CEUR}
  Workshop Proceedings, G.~Faggioli, N.~Ferro, A.~Joly, M.~Maistro, and
  F.~Piroi, Eds., vol. 2936.\hskip 1em plus 0.5em minus 0.4em\relax
  CEUR-WS.org, 2021, pp. 520--530. [Online]. Available:
  \url{http://ceur-ws.org/Vol-2936/paper-41.pdf}
\BIBentrySTDinterwordspacing

\bibitem{martin2021factercheck}
A.~Martín, J.~Huertas-Tato, Álvaro Huertas-García, G.~Villar-Rodríguez, and
  D.~Camacho, ``Facter-check: Semi-automated fact-checking through semantic
  similarity and natural language inference,'' 2021.

\bibitem{grootendorst2020bertopic}
M.~Grootendorst, ``Bertopic: Leveraging bert and c-tf-idf to create easily
  interpretable topics.'' 2020.

\bibitem{grootendorst2020keybert}
------, ``Keybert: Minimal keyword extraction with bert.'' 2020.

\bibitem{reimers_making_2020}
N.~Reimers and I.~Gurevych, ``Making monolingual sentence embeddings
  multilingual using knowledge distillation,'' 2020.

\bibitem{muller_introduction_2001}
K.-R. Muller, S.~Mika, G.~Ratsch, K.~Tsuda, and B.~Scholkopf, ``An introduction
  to kernel-based learning algorithms,'' \emph{IEEE Transactions on Neural
  Networks}, vol.~12, no.~2, pp. 181--201, Mar. 2001.

\bibitem{hyvarinen_independent_2013}
A.~Hyvärinen, ``\BIBforeignlanguage{en}{Independent component analysis: recent
  advances},'' \emph{\BIBforeignlanguage{en}{Philosophical Transactions of the
  Royal Society A: Mathematical, Physical and Engineering Sciences}}, vol. 371,
  no. 1984, p. 20110534, Feb. 2013.

\bibitem{scholkopf_nonlinear_1998}
B.~Schölkopf, A.~Smola, and K.-R. Müller, ``{Nonlinear Component Analysis as
  a Kernel Eigenvalue Problem},'' \emph{Neural Computation}, vol.~10, no.~5,
  pp. 1299--1319, 07 1998.

\bibitem{mcinnes2018umap-software}
L.~McInnes, J.~Healy, N.~Saul, and L.~Grossberger, ``Umap: Uniform manifold
  approximation and projection,'' \emph{The Journal of Open Source Software},
  vol.~3, no.~29, p. 861, 2018.

\bibitem{scikit-learn}
F.~Pedregosa, G.~Varoquaux, A.~Gramfort, V.~Michel, B.~Thirion, O.~Grisel,
  M.~Blondel, P.~Prettenhofer, R.~Weiss, V.~Dubourg, J.~Vanderplas, A.~Passos,
  D.~Cournapeau, M.~Brucher, M.~Perrot, and E.~Duchesnay, ``Scikit-learn:
  Machine learning in {P}ython,'' \emph{Journal of Machine Learning Research},
  vol.~12, pp. 2825--2830, 2011.

\bibitem{wolf-etal-2020-transformers}
\BIBentryALTinterwordspacing
T.~Wolf, L.~Debut, V.~Sanh, J.~Chaumond, C.~Delangue, A.~Moi, P.~Cistac,
  T.~Rault, R.~Louf, M.~Funtowicz, J.~Davison, S.~Shleifer, P.~von Platen,
  C.~Ma, Y.~Jernite, J.~Plu, C.~Xu, T.~L. Scao, S.~Gugger, M.~Drame, Q.~Lhoest,
  and A.~M. Rush, ``Transformers: State-of-the-art natural language
  processing,'' in \emph{Proceedings of the 2020 Conference on Empirical
  Methods in Natural Language Processing: System Demonstrations}.\hskip 1em
  plus 0.5em minus 0.4em\relax Online: Association for Computational
  Linguistics, Oct. 2020, pp. 38--45. [Online]. Available:
  \url{https://www.aclweb.org/anthology/2020.emnlp-demos.6}
\BIBentrySTDinterwordspacing

\bibitem{Sanh2019DistilBERTAD}
V.~Sanh, L.~Debut, J.~Chaumond, and T.~Wolf, ``Distilbert, a distilled version
  of bert: smaller, faster, cheaper and lighter,'' \emph{ArXiv}, vol.
  abs/1910.01108, 2019.

\bibitem{xlm-roberta-2019}
A.~Conneau, K.~Khandelwal, N.~Goyal, V.~Chaudhary, G.~Wenzek, F.~Guzm{\'{a}}n,
  E.~Grave, M.~Ott, L.~Zettlemoyer, and V.~Stoyanov, ``Unsupervised
  cross-lingual representation learning at scale,'' \emph{CoRR}, vol.
  abs/1911.02116, 2019.

\bibitem{liu2019roberta}
Y.~Liu, M.~Ott, N.~Goyal, J.~Du, M.~Joshi, D.~Chen, O.~Levy, M.~Lewis,
  L.~Zettlemoyer, and V.~Stoyanov, ``Roberta: {A} robustly optimized {BERT}
  pretraining approach,'' \emph{CoRR}, vol. abs/1907.11692, 2019.

\bibitem{feng2020languageagnostic}
F.~Feng, Y.~Yang, D.~Cer, N.~Arivazhagan, and W.~Wang, ``Language-agnostic bert
  sentence embedding,'' 2020.

\bibitem{reimers-etal-2016-task}
N.~Reimers, P.~Beyer, and I.~Gurevych, ``Task-oriented intrinsic evaluation of
  semantic textual similarity,'' in \emph{Proceedings of {COLING} 2016, the
  26th International Conference on Computational Linguistics: Technical
  Papers}.\hskip 1em plus 0.5em minus 0.4em\relax Osaka, Japan: The COLING 2016
  Organizing Committee, 2016, pp. 87--96.

\bibitem{wang-etal-2018-glue}
A.~Wang, A.~Singh, J.~Michael, F.~Hill, O.~Levy, and S.~Bowman, ``{GLUE}: A
  multi-task benchmark and analysis platform for natural language
  understanding,'' in \emph{Proceedings of the 2018 {EMNLP} Workshop
  {B}lackbox{NLP}: Analyzing and Interpreting Neural Networks for {NLP}}.\hskip
  1em plus 0.5em minus 0.4em\relax Brussels, Belgium: Association for
  Computational Linguistics, Nov. 2018, pp. 353--355.

\bibitem{incremental_pca_2008}
D.~A. Ross, J.~Lim, R.-S. Lin, and M.-H. Yang,
  ``\BIBforeignlanguage{eng}{Incremental learning for robust visual
  tracking},'' \emph{\BIBforeignlanguage{eng}{International journal of computer
  vision}}, vol.~77, no. 1-3, pp. 125--141, 2007.

\bibitem{bishop_2006}
C.~M. Bishop, \emph{Pattern Recognition and Machine Learning (Information
  Science and Statistics)}.\hskip 1em plus 0.5em minus 0.4em\relax Berlin,
  Heidelberg: Springer-Verlag, 2006.

\bibitem{ica_image_retrieva_2004}
C.~Liu, ``Enhanced independent component analysis and its application to
  content based face image retrieval,'' \emph{IEEE TRANSACTIONS ON SYSTEMS MAN
  AND CYBERNETICS PART B-CYBERNETICS}, vol.~{34}, no.~{2}, pp. {1117--1127},
  {APR} {2004}.

\bibitem{ica_face_recog_2005}
H.~Ekenel and L.~Sankur, ``Multiresolution face recognition,'' \emph{IMAGE AND
  VISION COMPUTING}, vol.~{23}, no.~{5}, pp. {469--477}, {MAY 1} {2005}.

\bibitem{mikolov2013efficient}
T.~Mikolov, K.~Chen, G.~Corrado, and J.~Dean, ``Efficient estimation of word
  representations in vector space,'' 2013.

\bibitem{umap_app_2019_nature}
J.~Cao, M.~Spielmann, X.~Qiu, X.~Huang, D.~M. Ibrahim, A.~J. Hill, F.~Zhang,
  S.~Mundlos, L.~Christiansen, F.~J. Steemers, C.~Trapnell, and J.~Shendure,
  ``The single-cell transcriptional landscape of mammalian organogenesis,''
  \emph{NATURE}, vol. {566}, no. {7745}, pp. {496+}, {FEB 28} {2019}.

\bibitem{umap_carter2019activation}
S.~Carter, Z.~Armstrong, L.~Schubert, I.~Johnson, and C.~Olah, ``Activation
  atlas,'' \emph{Distill}, 2019, https://distill.pub/2019/activation-atlas.

\end{thebibliography}

\end{document}